\pdfoutput=1

\documentclass[11pt]{article}

\usepackage{acl}

\usepackage{times}
\usepackage{latexsym}

\usepackage[T1]{fontenc}

\usepackage[utf8]{inputenc}

\usepackage{microtype}

\usepackage{inconsolata}

\usepackage{algorithm}
\usepackage{algorithmic}
\usepackage{amsmath}

\usepackage{hyperref}
\usepackage{url}
\usepackage{amssymb}

\usepackage{graphicx}
\usepackage{booktabs}       
\usepackage{amsfonts}       
\usepackage{nicefrac}       
\usepackage{microtype}
\usepackage{wrapfig}
\usepackage{graphicx}
\usepackage{multirow}
\usepackage{arydshln}
\usepackage{amsmath}
\usepackage{xspace}

\definecolor{mygreen}{rgb}{0.0, 0.5, 0.0}
\definecolor{myred}{rgb}{0.8, 0.0, 0.0}

\usepackage{amssymb} 
\usepackage{pifont} 

\newif\ifshowcomments

\showcommentstrue  

\definecolor{darkgreen}{RGB}{0,112,0}
\title{Instructions for *ACL Proceedings}

\author{Kenan Jiang$^*$$^{\dag}$\\
\normalsize UC Berkeley\\
{\tt\small kenanj11@berkeley.edu}
\And
Xuehai He$^*$\\
\normalsize UC Santa Cruz\\
{\tt\small xhe89@ucsc.edu}
\And
Ruize Xu$^{\dag}$\\
\normalsize Columbia University\\
{\tt\small rx2246@columbia.edu}
\And
Xin Eric Wang\\
\normalsize UC Santa Cruz\\
{\tt\small xwang366@ucsc.edu}
}

\begin{document}
\title{ComCLIP: Training-Free Compositional Image and Text Matching}

\maketitle
\def\thefootnote{*}\footnotetext{Equally contributed.}\def\thefootnote{\arabic{footnote}}
\def\thefootnote{\dag}\footnotetext{Work done during internship at UC Santa Cruz.}

\definecolor{darkgreen}{RGB}{0,112,0}

\begin{abstract}
Contrastive Language-Image Pretraining (CLIP) has demonstrated great zero-shot performance for matching images and text. However, it is still challenging to adapt vision-lanaguage pretrained models like CLIP to compositional image and text matching --- a more challenging image and text matching task requiring the model's understanding of compositional word concepts and visual components. Towards better compositional generalization in zero-shot image and text matching, in this paper, we study the problem from a causal perspective: the erroneous semantics of individual entities are essentially confounders that cause the matching failure. Therefore, we propose a novel \textbf{\textit{training-free}} compositional CLIP model (ComCLIP). ComCLIP disentangles input images into subjects, objects, and action subimages and composes CLIP's vision encoder and text encoder to perform evolving matching over compositional text embedding and subimage embeddings. In this way, ComCLIP can mitigate spurious correlations introduced by the pretrained CLIP models and dynamically evaluate the importance of each component. Experiments on four compositional image-text matching datasets: Winoground, VL-checklist, SVO, and ComVG, and two general image-text retrieval datasets: Flick30K, and MSCOCO demonstrate the effectiveness of our plug-and-play method, which boosts the \textbf{\textit{zero-shot}} inference ability of CLIP, SLIP, and BLIP2 even without further training or fine-tuning. 
Our codes can be found at \url{https://github.com/eric-ai-lab/ComCLIP}.
\end{abstract}

\section{Introduction} \label{sec:label}

\begin{figure}[t]
    \centering
    \includegraphics[width=0.5\textwidth]{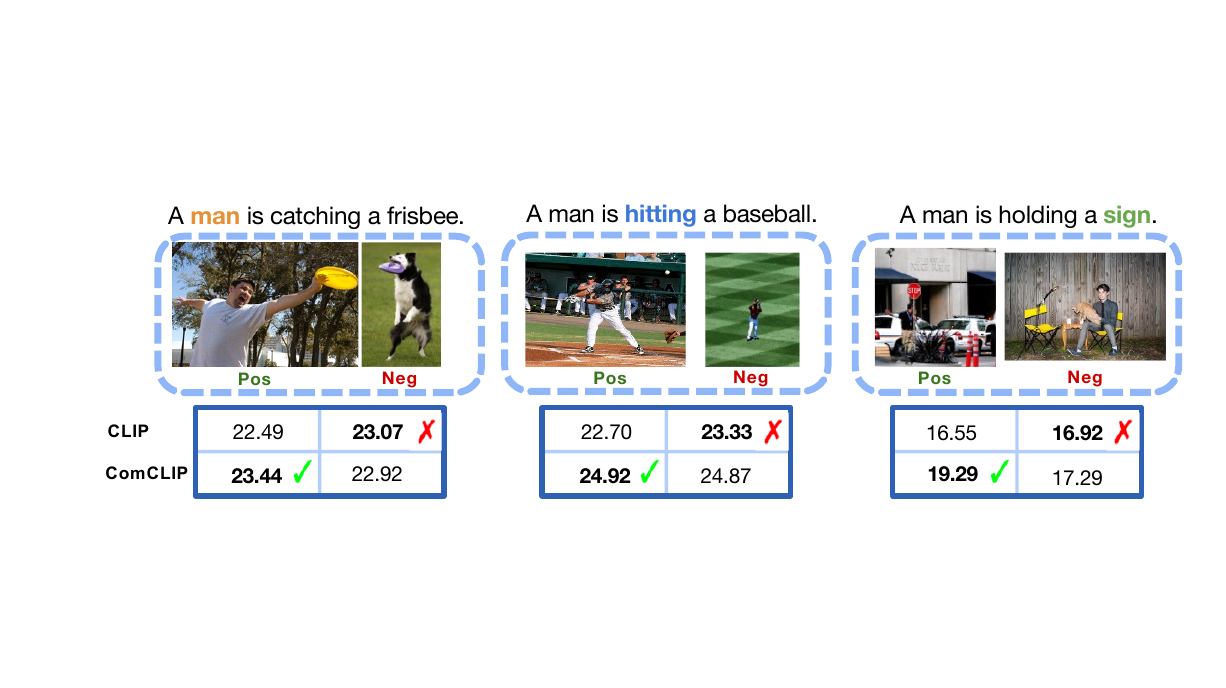}
    \caption{Examples of the compositional image-text matching problem, in which the positive and negative images have very similar semantics except for the only difference in subject, predicate/verb, or object. CLIP mistakenly connects the text prompts with the wrong images on the right (high similarity scores with negative images), while our ComCLIP model does compositional matching more effectively. 
    }
    \label{fig:teaser}
      \vspace{-1ex}
\end{figure}
Image and text matching~\cite{flickr30k, coco} is a fundamental task for vision-language research that involves multimodal reasoning and multi-level visual and text concept alignment.
Recently, a growing number of pretrained vision-language foundation models~\cite{clip,align,blip, GLIP} have shown encouraging results towards open-domain visual and language concept matching. 
Among these models, CLIP~\cite{clip} can be easily transferred to image and text matching under zero-shot and few-shot scenarios. 
However, CLIP treats the image and the text as a whole for alignment and ignores the compositional matching of disentangled concepts, especially for tasks that require the model's compositional understanding ability.
For instance, Figure~\ref{fig:teaser} shows some examples that CLIP fails at, which require a compositional generalization of the model to understand different subject, predicate, or object concepts.

In fact, it is widely observed that current pretrained vision-language models struggle to recognize actions from the image, distinguishing objects from subjects~\cite{SVO_dataset}, or failing to identify objects in unseen surroundings~\cite{elephant}. 
These may be ascribed to shortcut learning~\cite{shortcut} and dataset biases in pretraining, where the models learn the correspondence between entities and images implicitly and are thus vulnerable to spurious correlations, incurring biases toward particular objects/subjects/predicates and combinations. 

Therefore, there are primarily two challenges to address when adopting CLIP for compositional image and text matching.  \emph{Challenge 1}: the pretrained language model in CLIP is biased and tends to rely on spurious relationships learned in pretraining. For example, in Figure~\ref{fig:teaser} (A), CLIP associates ``frisbee'' with ``dog'' because of their more frequent co-occurrence and makes the wrong prediction.  Meanwhile, the richness of entities in text descriptions brings \emph{Challenge 2}: entity embeddings should contribute dynamically for compositional matching. In Figure~\ref{fig:teaser}, the subject/predicate/object entities ``{\color{orange}man}/{\color{blue}hitting}/{\color{teal}sign}'', as identifiers for correct matching in each scenario, should be endowed with more importance. Based on the semantics of input images, CLIP should adjust the weights for these entity embeddings. Yet existing approaches often calculate the similarities merely based on the global embedding of images and texts and overlook fine-grained concept matching~\cite{Visual_semantic_reasoning_for_image_text_matching}.

To address the above limitations, we propose a new \textit{\textbf{training-free}} framework based on CLIP-like models from the causal viewpoint, named ComCLIP. Specifically, we disentangle the visual scene into individual visual concepts and construct counterfactual subimages containing subject/object/predicate entities only. Then we utilize backdoor adjustment~\cite{backdoor_adjustment} to implement interventions over the disentangled subimages to mitigate the effect of spurious correlations. 
With this design, ComCLIP can bind the disentangled visual components with the correct word concept and avoid matching solely based on spurious correlations learned during pretraining and fine-tuning, achieving compositional generalization. 
To validate our approach, we formalize the~\underline{compositional image and text matching} task and construct a new Compositional Visual Genome (ComVG) dataset from the Visual Genome~\cite{visualgenome} dataset for this task. We evaluated on multiple datasets: Winoground, VL-checklist, SVO-Probes~\cite{SVO_dataset}, Flickr30K~\cite{flickr30k}, MSCOCO~\cite{coco}, and the ComVG dataset. Notably, ComCLIP gains an absolute accuracy improvement of 4.50\% on the image score and 2.34\% on the group score over CLIP and SLIP respectively on the challenging Winoground dataset.

Our contributions are summarized as follows:
\begin{itemize}
    \item We formally define the compositional image and text matching problem and propose a novel approach named ComCLIP to address it from the causal perspective: disentangling the input image into counterfactual subimages and leverages the backdoor adjustment~\cite{backdoor_adjustment} to compose entity features and perform fine-grained compositional concept matching, mitigating the effect of spurious correlations introduced during training and achieving compositional generalization. 
    \item The ComCLIP framework is \textbf{\textit{training-free}} and can be applied to CLIP-like models for \textbf{\textit{zero-shot}} inference without further training.
    \item We introduce a new dataset, the Compositional Visual Genome\footnote{The dataset is available at 
\href{https://drive.google.com/file/d/1rWHuq48paToXZs7_OT2Wko4l5YrAfFmR/}{
{https://drive.google.com/file/d/1rWHuq48pa}\\{ToXZs7\_OT2Wko4l5Y}
{rAfFmR/view}}}
, which contains 5400 image-text pairs with \textbf{s}ubject, \textbf{v}erb, and \textbf{o}bject annotations. This dataset was generated by creating image–sentence pairs from the Visual Genome~\cite{visualgenome} in the same format as the SVO-Probes~\cite{SVO_dataset} dataset, to benchmark compositional image and text matching.
    \item We demonstrate the effectiveness of ComCLIP by outperforming CLIP on the Winoground, VL-checklist, SVO-Probes, and ComVG dataset over the compositional image-text matching task. We also shows its effectiveness over the general image-text retrieval task by testing Flickr30K and MSCOCO.

\end{itemize}

\begin{figure*}[t]
    \centering
    \includegraphics[width=\textwidth]{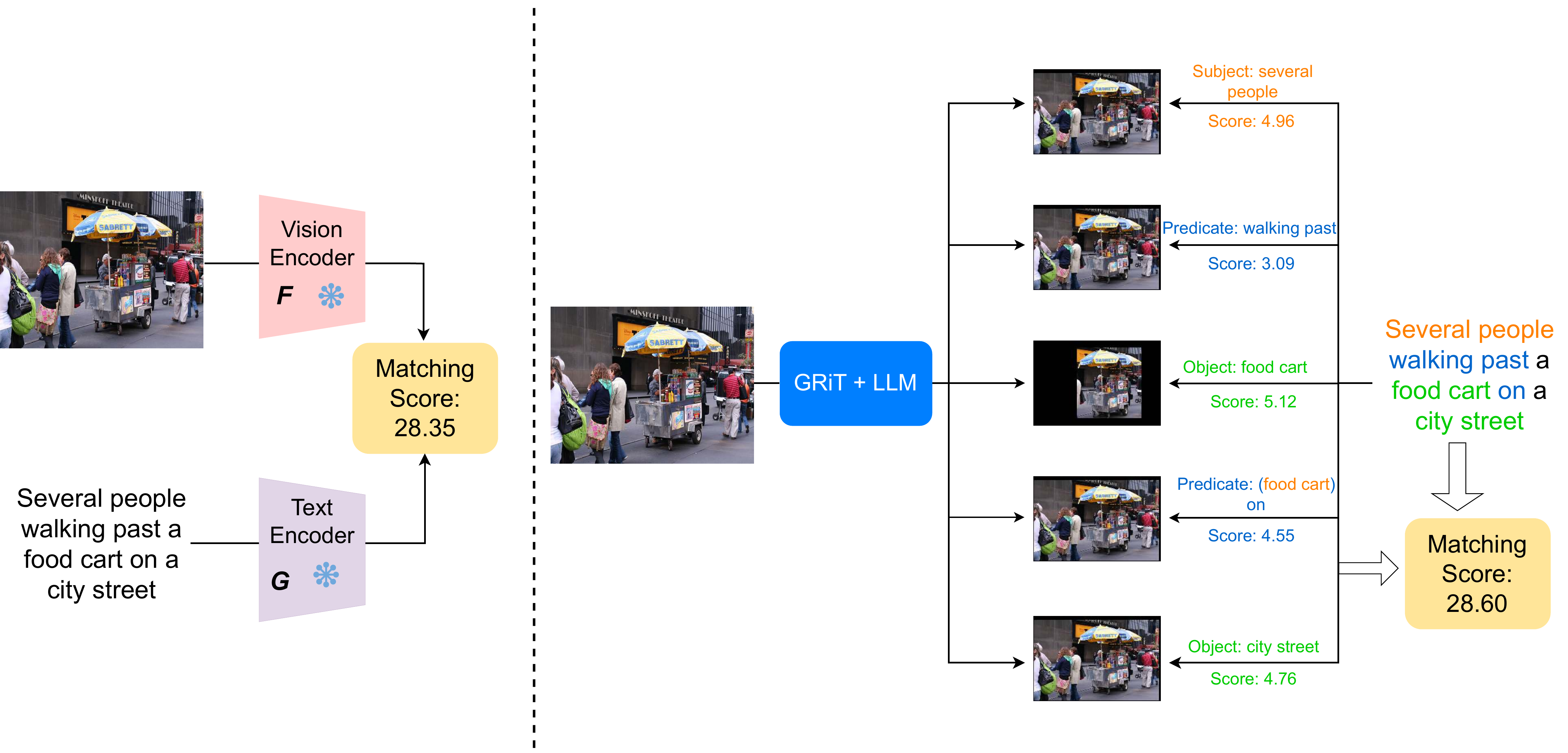}
    \caption{Overview of our ComCLIP framework using CLIP as the backbone. We disentangle the input image using GRiT~\cite{wu2022grit} and the Large Language Model (LLM) by obeying the rules of encoding object, subject, and predicate respectively. 
    The figure shows the case where multiple subjects/objects/predicates are involved (this is a positive example from Flickr30K).
    }
    \label{case_study1}
\end{figure*}

\begin{figure*}[t]
    \centering
    \includegraphics[width=\textwidth]{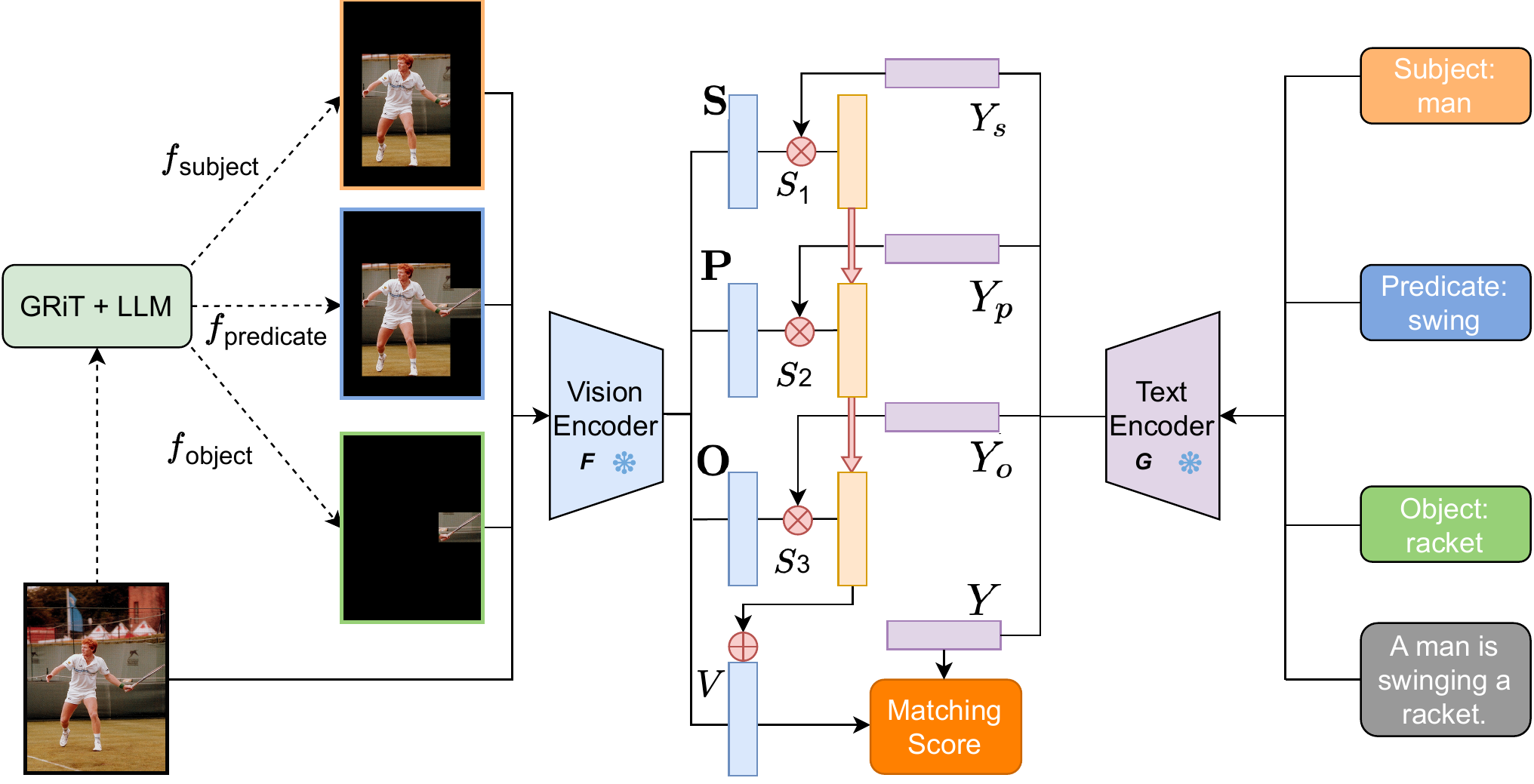}
    \caption{Overview of our ComCLIP framework using CLIP as the backbone. We disentangle the input image using three independent encoding mechanisms by obeying the rules of encoding object, subject, and predicate respectively. The entity information is introduced to the global embedding of the whole image. Module components from CLIP (vision encoder $F(\cdot)$, text encoder $G(\cdot)$) are always frozen. During implementation, the process for matching and calculating the score begins with the input image being processed into object, subject, and predicate sub-images. This is followed by feeding both the original sentence and image, along with their parsed words and sub-images, into the CLIP text and vision encoders. Subsequently, cosine similarity scores are computed for each pairing of sub-image and word embeddings. These scores are then subjected to a Softmax layer, resulting in three positive weights. The next step involves adding the reweighted sub-image embeddings to the embedding of the original image. Finally, the ultimate matching score is derived from comparing this aggregated image embedding and the global text embedding.The whole framework is \textbf{\textit{training-free}}.
    }
    \label{fig:overview}
      \vspace{-1ex}
\end{figure*}

\section{Related Work} \label{sec:related}
\noindent\textbf{Image-Text Matching}
Most existing image-text matching datasets are evaluated in a classification setting. For example, ~\cite{hico,Visual_relationship_detection} focus on the relationship or interaction detection.
~\cite{contrastive_weakly_phrase_grounding,faghri2017vse++} explore how creating hard negatives (e.g., by substituting words in train examples) leads to better test performance. FOIL benchmark~\cite{shekhar2017foil} tests if vision-language models can differentiate between sentences that vary with respect to only one noun. SVO-Probes adds hard evaluation examples to test the model's understanding of verbs as well as subjects and objects in a controlled way. To associate local regions in an image with texts to do matching,  ~\cite{xu2015show} incorporates a soft form of attention into their recurrent model.
~\cite{ma2015multimodal} learns multiple networks that capture words, phrases, and sentence-level interactions with images and combines the scores of these networks to obtain a whole image-sentence score.  ~\cite{hu2016natural} leverages spatial information and global context to predict where objects are likely to occur. ~\cite{wang2016structured} formulates a linear program to localize all the phrases from a caption jointly. In this paper, we focus on the task of matching error-prone texts with images, requiring distinguishing words on a granular level --- compositional image and text matching.

\noindent\textbf{Pretrained Vision-Language Models}
Vision-Language models pretrained on large-scale image-text pairs have demonstrated great potential in multimodal representation  learning~\cite{align,flip,florence,GLIP,clip}. Among them, CLIP~\cite{clip} benefits from 400M curated data and defines various prompt templates to carry out zero-shot image classification. GLIP~\cite{GLIP} has incorporated region-level alignment in its pretraining. 
However, these models can suffer from connecting verbs/subjects/objects concepts with visual components correctly~\cite{SVO_dataset} and bias towards spurious relations they have seen in the pretraining data, referred to as ``confounders" ~\cite{devlbert}.
By modeling using a structural causal model (SCM) network~\cite{structural_causal_models}, ~\cite{devlbert} executes a hard intervention to eliminate dataset bias via a backdoor intervention during pretraining. Different from them, in this work, we focus on mitigating the effect of spurious relations and improving the zero-shot inference and compositonal generalization abilities of off-the-shelf pretrained vision-language models. We develop a new training-free paradigm that gains superior performance on compositional image and text matching.

\noindent\textbf{Disentangled Representation Learning}
It is often assumed that real-world observations like images can be disentangled~\cite{representation_learning, elements_of_causal_inference}. \cite{disentangle_image_generation} disentangles background, texture, shape, etc., and uses object bounding boxes as supervision to synthesize images. 
~\cite{besserve2020counterfactuals} leverages the idea of independent mechanisms to identify modularity in pretrained generative models. ~\cite{personReID_align} performs hierarchical alignments in three different granularities, i.e., global-global, global-local, and local-local alignments for description-based person
re-id. ~\cite{fine_video_text_retrieval} improves fine-grained video-text retrieval by decomposing video-text matching into global-to-local levels. 
~\cite{zhang2022multi} proposes a multi-granularity semantic collaborative reasoning network and employs different granularity semantic representations of the question and dialog history to collaboratively identify the relevant information from multiple inputs based on attention mechanisms. ~\cite{counterfactual_generative_networks} utilizes independent mechanisms to generate images to improve image classification. ~\cite{eiclip} disentangles word entities from the conventional meanings of special entities encoded in the pretrained language model. None of these works consider the alignment of subjects, objects, and predicate entities. Different from them~\cite{elements_of_causal_inference}, we employ independent mechanisms to disentangle images and use generated subimages to improve fine-grained visual and language concept matching, which can mitigate spurious correlations introduced by the pretrained model.

\section{Compositional Image and Text Matching}
We first introduce the task of compositional image and text matching, where we are interested in improving the compositional understanding, more specifically, subject/object/predicate understanding of vision-language models. Compositional image and text matching is a task focused on enhancing the understanding of compositional elements such as subjects, objects, and predicates within CLIP-like models. This task requires an appreciation of fine distinctions between texts and their underlying compositional structure, as illustrated in Figure~\ref{fig:teaser} with phrases like ``{\color{orange}man}/{\color{blue}hitting}/{\color{teal}sign}." The model's ability to differentiate images that only vary by one conceptual element in their accompanying text highlights its comprehension of compositionality.

We formally define this task as follows: given text prompts $Y$ (e.g., 
"\texttt{A man is hitting a baseball}") and a set of entities $T^E=\{e^k\}_{k=1}^K$ such as {\color{blue}hitting}, where $K$ denotes the total number of entities and $e^k$ represents the $k$-th entity, the model's objective is to match the text prompts with the corresponding images. The challenge lies in the inclusion of negative images that contain mismatched entities $\{e^k\}_{k=1}^n$, where $n<k$. These negative images are designed to confuse the model, demanding a nuanced understanding of the entities within a sentence. Simply relying on nouns or spurious relations would not succeed at this task.
To evaluate how well the model grasps this concept of compositionality in texts and matches them with the right images, we introduce an additional ComVG dataset as an extended testing platform.

\section{ComCLIP} \label{sec:method}
We propose ComCLIP to incorporate a causal view into the CLIP-like models. We briefly introduce the background of ComCLIP in view of structured causal models in Section~\ref{sec:background}. Then, we present the overview of ComCLIP pipeline in Section~\ref{subsec:m1}. We introduce its critical components in depth in Section~\ref{visual_concept} and~\ref{composition}. Our objectives are: (i) We aim at disentangling visual input into subimages containing fine-grained compositional concepts. (ii) We intend to utilize those disentangled concepts to perform entity-level matching dynamically and mitigate the effect of spurious relations in the pretrained vision-language models learned during training. 

\subsection{Background}
\label{sec:background}
Causal inference aims to understand how changing one variable can affect another, often represented using concepts such as confounders, interventions, counterfactuals, and do-operations. In the realm of computer vision and natural language processing, the causal relationships can provide insights into the underlying generative processes.

Consider a dataset comprised of (high-dimensional) observations $X$ (i.e., images) and corresponding text prompts $Y$. Assume that each $X$ can be described by lower-dimensional, semantically meaningful factors of variation $z$ (e.g., objects, subjects, or action relations between objects and subjects (i.e., predicates in the image)). These factors, which we term confounders $Z$, may affect either $X$ or $Y$. By disentangling these factors, we can achieve more granular image and text matching. This idea of disentanglement resonates with the principles of structural causal models (SCMs)~\cite{structural_causal_models} and independent mechanisms (IMs). An SCM is a mathematical formulation representing how variables influence one another, often composed of multiple IMs, the individual causal processes. Inspired by SCMs, our approach decomposes the subimage generation process into three independent mechanisms: object mechanism $f_{\text {object }}$, subject mechanism $f_{\text {subject}}$, and predicate mechanism $f_{\text {predicate}}$.

\subsection{Method Overview}
\label{subsec:m1}
We introduce the overview of our method from a conceptual view. The pipeline is shown in Figure~\ref{case_study1} and Figure~\ref{fig:overview}. 
Our goal is to refine a pretrained vision-language model for fine-grained compositional image-text matching. This involves disentangling an input image to create entity-specific subimages, calculating similarity scores between these subimages and their textual counterparts, and integrating these weighted embeddings with the global image embedding. This process enables the model to capture non-spurious semantic entity information and conduct concept matching at the granular level.

\subsection{Counterfactual Subimage Generation}
\label{visual_concept}
Our method centers on the concept of causality, particularly, the Independent Mechanism (IM) assumption. In the realm of causality, the IM assumption posits that a system's variable generation process comprises autonomous modules that operate without mutual interference~\cite{elements_of_causal_inference}. We adopt this principle and tailor it to our context by considering three independent mechanisms for generating object, subject, and predicate subimages.

While our method is inspired by causal mechanisms, we do not make strong causal claims. Instead, we utilize the intuition that in a complex system, certain variables (or mechanisms) operate autonomously. Given the aforementioned setup, our structural causal model (SCM) takes the form:
$
\mathbf{O} :=f_{\text {object }}\left(X\right), \mathbf{S} :=f_{\text {subject}}\left(X\right), \mathbf{P} :=f_{\text {predicate}}\left(X\right).
$
Where $\mathbf{O}$ is the object image, $\mathbf{S}$ is the subject image, and $\mathbf{P}$ is the predicate image. 


With the structural framework above, we answer counterfactual questions, a fundamental concept in causality. Specifically, we pose questions like "What if we retain only the subject/object/predicate in the original image?". The responses to such inquiries allow us to generate what we term as \emph{counterfactual subimages}. The essence of these images is that they exclusively feature the entity in question (see Figure~\ref{fig:overview}). This procedure leads to the disentanglement of the input image into three distinct and causally independent subimages.

With these foundational blocks in place, our method is geared to connect each disentangled image entity with its corresponding textual counterpart. When each entity is independently and aptly encoded, matching becomes streamlined and efficient. The remaining challenge is to craft a mechanism that effectively governs the composition process of distinct entity regions within an image.

\subsection{Entity Composition}
\label{composition}
As mentioned, the pretrained CLIP-like model is prone to be biased toward specific subjects, objects or predicates, or even rely solely on one of them in the sentence.

From the causal perspective, to match image $X$ with text prompt $Y$ correctly, we want to infer $P(Y|X)$ while at the same time mitigating the effect of detrimental confounders $z$. The confounders may introduce spurious correlations in the model when directly inferring from $P(Y \mid X)$. 

Our goal is to infer $P(Y \mid X)$ while mitigating the effects of detrimental confounders $z$.
Leveraging Bayes Rule, 
\begin{align}
P(Y \mid X)=&\sum_z P(Y, z \mid X)
\\ =&\sum_z P(Y \mid X, z) {P(z \mid X)},
\end{align}
the confounder $z$ introduces the bias of word concept via $P(z \mid X)$. To adjust the effect of confounder $z$, we can intervene $X$ by first disentangling it and then intervening with it using $do$-operation~\footnote{$P(Y \mid do(X)$ uses the do-operator~\cite{do-operator}. Given random variables $X, Y$, we write $P(Y= y \mid d o(X=x))$ to indicate the probability that $Y=y$ when we intervene and set $X$ to be $x$. }: \begin{equation}
P(Y \mid do(X))=\sum P(Y \mid X, z){P(z)}.
\end{equation}
$do(X)$ refers to the process of mitigating the effect of harmful confounders $z$. These confounders $z$, as explained in Section 4.1, are lower-dimensional and semantically meaningful factors that include objects, subjects, and predicates within the image. By mitigating the impact of these confounders, we aim to refine our compositional matching process between the image and text.
We now seek an implicit way to compute $P(Y \mid X, z)$ and ${P(z)}$. Considering the SCMs mentioned above,
we interpret  $f_{\text {object }}(X), f_{\text {subject}}(X), f_{\text {predicate}}(X)$ as incorporating entity semantics into attended regions. 

To do concept matching over the text prompt $Y$ and the entity set $T^E=\left\{e^k\right\}_{k=1}^K$, where $K$ is the total number of entities, and $e^k$ is the $k$-th entity. $T^E$ represents a set of entities extracted from text prompts, during testing, both the image and its corresponding text, along with these parsed entities and their associated subimages, are processed through the CLIP text and vision encoders.

This interpretation motivates us to compute the similarity between $f_{\text {object }}(X), f_{\text {subject}}(X), f_{\text {predicate}}(X)$ with different word entity embeddings to achieve concept-wise semantic fusion and guidance. The prediction $P(Y \mid X, z)$ can be regarded as a classifier: $P(Y \mid X, z)=$ Softmax $f_i(X, z)$. Similar to~\cite{wangVisualCommonsenseRCNN2020}, using the approximation of NGSM (Normalized Weighted Geometric Mean)~\cite{show_attend_and_tell}, we have:
$
P(Y \mid do(X)) \approx \operatorname{Softmax}\left[\mathbb{E}_z\left(f_i(X, z)\right)\right].
$
Specifically, to implement this on the ComVG dataset, given an input image $X$ and IMs $f_{\text {object }}(\cdot), f_{\text {subject}}(\cdot), f_{\text {predicate}}(\cdot)$, we first extract a collection of visual concepts from input images as $f_{\text {object }}(X), f_{\text {subject}}(X), f_{\text {predicate}}(X)$. For the language side, given a prompt $Y$ and its entity set $T^E$, we extract all (subject, object, predicate) words ($Y_s, Y_o, Y_p$) from the input text prompts. Using cosine similarity score $\mathcal{S}$ as an example, we compute the concept-level similarity separately:
\begin{equation}
\begin{aligned}
    & S_1 =   \mathcal{S}(F(f_{\text {object }}(X)), G(Y_s)), \\& S_2 = \mathcal{S}(F(f_{\text {subject }}(X)), G(Y_o)), \\& S_3=S(F(f_{\text {predicate}}(X)), G(Y_p)),\; \\& \text{where} \ F(\cdot)=\text{CLIP}_{\text{vision}}(\cdot),\; G(\cdot) = \text{CLIP}_{\text{text}}(\cdot).
    \label{similarity}
\end{aligned}
\end{equation}
The final visual feature is composed by:
\begin{equation}
\begin{aligned}
   V = F(X) +& F(f_{\text {object }}(X))S_1  + F(f_{\text {subject }}(X))S_2  \\&+  F(f_{\text {predicate}}(X))S_3.\; 
   \label{eq:visual_feature}
\end{aligned}
\end{equation}
By adding compositional features back to the global image feature (as in Eq~\ref{eq:visual_feature}) and matching them with the global text features, we balance the need for detailed matching with overall context preservation.

We can compute the image-text matching score by:
$
    O = S(G(Y), V).
 $
With this design, the language part of CLIP is aware of connections between entities from both the visual and language input when doing the concept matching. During implementation, we calculate cosine similarity scores for each pair of subimage and word embedding. These scores are then transformed into weights using a Softmax layer. Subsequently, we enhance the original image embedding by adding these reweighted subimage embeddings. The final step involves computing the overall matching score by comparing this augmented image embedding with the global text embedding, thus finalizing our image-text matching process.

Our algorithm is summarized in Algorithm~\ref{algo:algo} in the Appendix, which requires \textit{\textbf{no training or additional data}}. Note that apart from CLIP, it can be easily adapted to other vision-language pretrained model with the two-stream encoder structure.

\section{Experiments} \label{sec:exp}
\subsection{Datasets}
\noindent\textbf{Winoground~\cite{thrush2022winoground}} Designed to evaluate vision-language models, this dataset contains 400 instances with two image-text pairs per instance. The challenge is the differing arrangement of identical words across the pairs. Our evaluation spanned the entire dataset.

\noindent\textbf{VL-checklist~\cite{vlchecklist}} Distinguishing itself by combining multiple sources, VL-checklist classifies 410,000 images into three categories. We analyzed a subset of 2000 images from each category to gauge our method's effectiveness.

\noindent\textbf{Flickr30K~\cite{flickr30k}} Each of the 1000 test images has 5 annotations; one annotation is selected randomly. CLIP is evaluated across the dataset; for ComCLIP, the top 10 similar images from CLIP are taken. We create subimages for the top 10 similar images and apply ComCLIP to them.

\noindent\textbf{MSCOCO~\cite{coco}} Like Flickr30K, for each of the 1000 test images, one annotation is selected randomly. The top 10 images from CLIP undergo ComCLIP processing, and subimages are created based on parsed elements.

\noindent\textbf{SVO-Probes~\cite{SVO_dataset}} Built to assess language-image models on distinctions within image elements. From its initial 30,000 data points, we utilized 13,000 due to accessibility issues. We conducted tests using three random divisions and presented the average accuracy.

\noindent\textbf{Compositional Visual Genome (ComVG)~} 
Derived from Visual Genome's~\cite{visualgenome} 2.3 million relationships, we developed ComVG. These relationships, encompassing action and spatial aspects, are in subject-predicate-object triplets. Using these, we created image descriptions and selected 542 distinct relationship images from Visual Genome. Similar to SVO-Probes, we identified variants for each image with single discrepancies in subject, object, or predicate, resulting in 5400 curated test samples with grammatical corrections. ComVG stands out for its high-quality images and focus on text-to-image retrieval.
For comprehensive dataset statistics, kindly refer Table~\ref{tab:main1}. Our evaluation covered the entire ComVG.

More data examples are presented in Appendix.

\begin{table}[t]  
\caption{
  The number of data samples in the dataset that have one of their subjects, objects, or predicates changed between positive and negative images and the number of unique types of subjects, predicates, and objects across ComVG and SVO-Probes (SVO).
  }
 \resizebox{\linewidth}{!}{
  \centering
  \begin{tabular}{ccccccc}
    \toprule
    & Sub-Neg & Pred-Neg &Obj-Neg  &    Subjects &
   Predicates  & Objects   \\
     \midrule
ComVG & 2,584 & 1,536&1,280 &   30 &
65 & 82  \\
  SVO & 5,679 & 23,525&7,637 &    100 &
  421 & 275   \\
\bottomrule
  \end{tabular}}
 \label{tab:main1}
  \end{table}

\begin{figure}[t]
    \centering
    \includegraphics[width=0.45\textwidth]{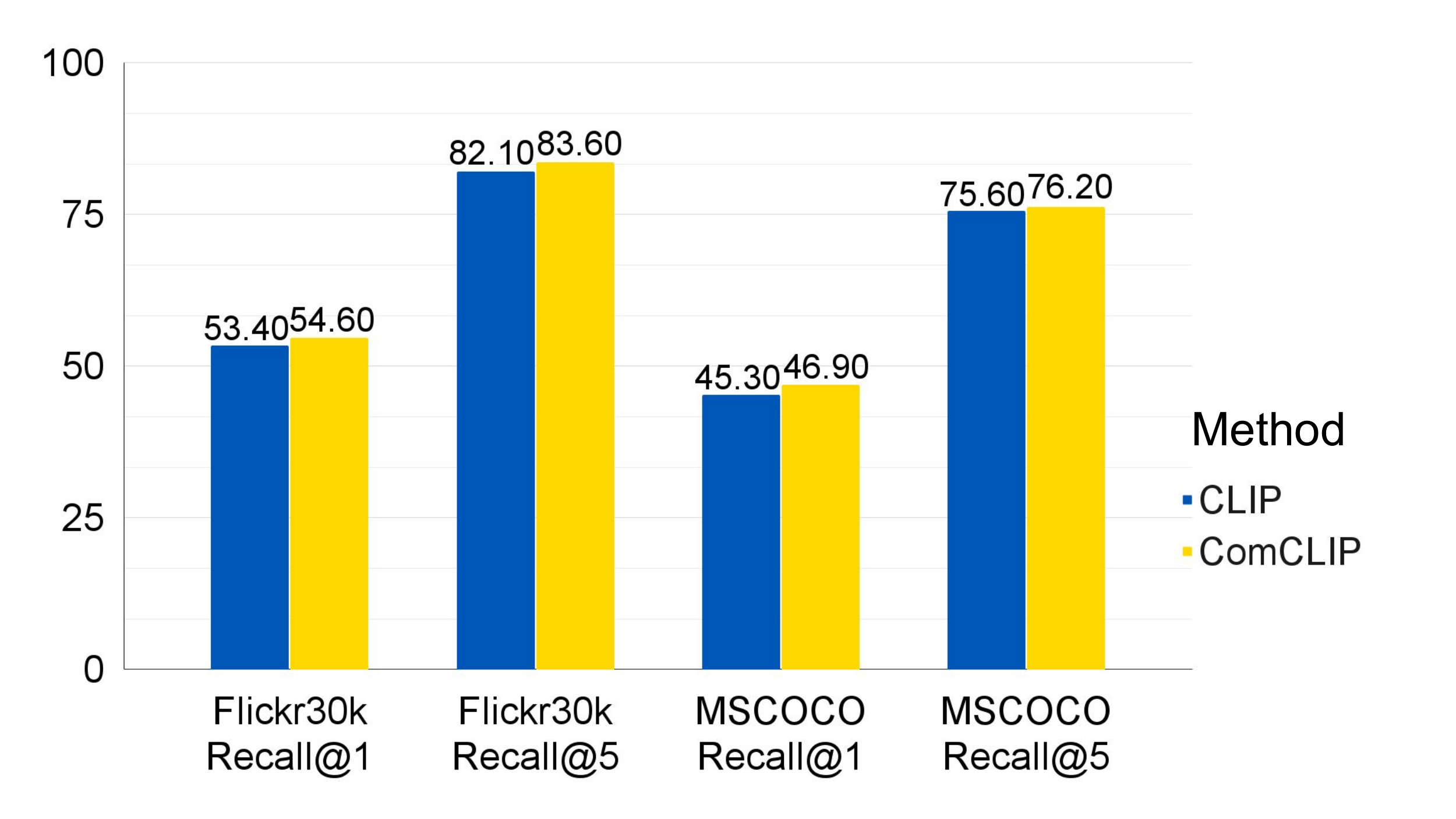}
    \caption{Comparison of Recall@1 (\%) and Recall@5 (\%) using CLIP and ComCLIP over the general image-text retrieval datasets. 
    }
    \label{fig:Flickr8K}
\end{figure}

\begin{table}[t]
  \caption{Comparison of accuracy (\%) on Winoground and VL-checklist using SLIP, and CLIP, and BLIP2. Results marked with $\spadesuit$ are our methods.}
  \centering
  \resizebox{\columnwidth}{!}{
    \begin{tabular}{llllllll}
      \toprule
      \multirow{2}*{Method} & \multicolumn{3}{c}{Winoground} & \multicolumn{4}{c}{VL-checklist} \\
      \cmidrule(lr){2-4} \cmidrule(lr){5-8}
      & {Text} & {Image} & {Group} & {Attribute} & {Object} & {Relation} & {Ave} \\
      \midrule 
      SLIP &  23.25 &  10.00 & 6.75  &  65.95 &  76.81 & 65.30 & 69.35  \\
      ComSLIP $\spadesuit$ & 26.76 \tiny $\mathcolor{red}{(+3.51)}$ &  12.12 \tiny $\mathcolor{red}{(+2.12)}$ & 9.09 \tiny $\mathcolor{red}{(+2.34)}$ & 67.64 \tiny $\mathcolor{red}{(+1.69)}$ & 77.79 \tiny $\mathcolor{red}{(+0.98)}$ & 67.02 \tiny $\mathcolor{red}{(+1.72)}$ & 70.82 \tiny $\mathcolor{red}{(+1.47)}$ \\
     \hdashline
      CLIP & 31.25 & 11.25 & 9.00 & 67.85 & 75.70 & 67.15 & 70.23 \\
      ComCLIP $\spadesuit$ & \textbf{34.00} \tiny $\mathcolor{red}{(+2.75)}$ & \textbf{15.75} \tiny $\mathcolor{red}{(+4.50)}$ & \textbf{10.50} \tiny $\mathcolor{red}{(+1.50)}$ & 69.90 \tiny $\mathcolor{red}{(+2.05)}$ & 79.00 \tiny $\mathcolor{red}{(+3.30)}$ & 69.30 \tiny $\mathcolor{red}{(+2.15)}$ & 72.73 \tiny $\mathcolor{red}{(+2.50)}$ \\
      \hdashline
      BLIP & 29.25 & 12.00 & 8.75 & 79.00 & 84.05 & 73.55 & 78.87 \\
      ComBLIP $\spadesuit$ & 28.75 \tiny $\mathcolor{red}{(-0.50)}$ & 13.00 \tiny $\mathcolor{red}{(+1.00)}$ & 10.00 \tiny $\mathcolor{red}{(+1.25)}$ & \textbf{79.15} \tiny $\mathcolor{red}{(+0.15)}$ & \textbf{84.70} \tiny $\mathcolor{red}{(+0.65)}$ & \textbf{73.95} \tiny $\mathcolor{red}{(+0.40)}$ & \textbf{79.27} \tiny $\mathcolor{red}{(+0.40)}$ \\
      \bottomrule
    \end{tabular}
  }
    \vspace{-1ex}
  \label{tab:winoground}
\end{table}

\begin{table}[t]
  \caption{Comparison of accuracy (\%) on ComVG, and average accuracy (\%) across the three splits on SVO-Probes using  CLIP, GLIP, and ComCLIP. Results marked with $\spadesuit$ are our methods. Ours could also beat GLIP, showing the superiority of our method compared with region-based vision-language pretrained models. 
  }
  \centering
 \resizebox{\columnwidth}{!}{
  \begin{tabular}{lcccccccc}
    \toprule
    \multirow{2}*{Method} & \multicolumn{4}{c}{ ComVG} & \multicolumn{4}{c}{SVO-Probes}\\
    \cmidrule(lr){2-5} \cmidrule(lr){6-9}
    & {Sub} &{Pred} &{Obj}& Ave&{Sub} &{Pred} &{Obj}&Ave \\
       \midrule 
        GLIP & 65.95 & 57.50 & 65.75 &63.85 & 68.91 & 65.14 & 74.94 &67.81\\ \hdashline
      SLIP & 86.20 & 61.33 & 85.84 &80.13 & 79.62 & 79.92 & 78.43 &79.57\\
ComSLIP $\spadesuit$ & 87.43 & 61.25 & 87.11 &81.07 & 79.73 & 80.83 & 79.63 &80.42\\ \hdashline
   CLIP &88.61 & 68.52 & 93.85&86.38 & 85.53 & 80.77 & 90.53 &85.60\\
   ComCLIP $\spadesuit$ &\textbf{90.04} & \textbf{69.06} & \textbf{94.78} &\textbf{87.40} & \textbf{86.70} & \textbf{81.87} & \textbf{90.67} &\textbf{86.41}\\
  \bottomrule
  \end{tabular}
  }
    \vspace{-1ex}
 \label{tab:main}
\end{table}

\begin{table}[t]
  \caption{Comparison of accuracy (\%) on Compositional Visual Genome and SVO-Probes using CLIP, OpenCLIP, and ComCLIP.}
  \vspace{-1ex}
 \resizebox{\columnwidth}{!}{
  \centering
  \begin{tabular}{lcccccccc}
    \toprule
    & \multicolumn{3}{c}{Compositional Visual Genome} & \multicolumn{3}{c}{SVO-Probes} \\
    \cmidrule(lr){2-4}\cmidrule(lr){5-7}
    Vision Encoder & {CLIP} & {OpenCLIP} & {ComCLIP} & {CLIP} & {OpenCLIP} & {ComCLIP} \\
  \midrule
   ResNet-50 &82.25 & 82.21 & \textbf{83.73} & 83.07 & 83.06 & \textbf{84.17} \\
   ViT-B-32 &82.45 & 82.41 & \textbf{84.75} & 84.28 & 84.27 & \textbf{85.18} \\
   ViT-L-14 &86.38 & 86.38 & \textbf{87.40} & 85.61 & 85.60 & \textbf{86.41} \\
  \bottomrule
  \end{tabular}
  }
 \label{tab:mergedResults}
\end{table}

\begin{table}[t]
  \caption{Comparison of accuracy (\%) on Compostional Visual Genome and SVO-Probes using different subimage configuration.
  }
    \vspace{-1ex}
 \setlength{\tabcolsep}{1pt}
 \resizebox{\columnwidth}{!}{
  \centering
  \begin{tabular}{lcccccc}
    \toprule
    & \multicolumn {3}{c}{Compositional Visual Genome} & \multicolumn{3}{c}{SVO-Probes} \\
    \cmidrule(lr){2-4}\cmidrule(lr){5-7}
    {Subimage Configuraion} & {ResNet-50} & {ViT-B-32} & {ViT-L-14} & {ResNet-50} & {ViT-B-32} & {ViT-L-14} \\
  \midrule
   ComCLIP & \textbf{83.73} & \textbf{84.73} & \textbf{87.40} & \textbf{84.17} & \textbf{85.18} & \textbf{86.41} \\
   All black subimages & 82.75 & 83.33 & 86.35&83.09 & 83.83 & 84.47\\
   All original images & 82.25 & 82.45 & 86.38 & 83.07 &84.27 & 85.60\\
   All subject subimages & 82.46 & 82.55 & 86.46 & 83.18 & 84.10&85.24 \\
   All object subimages & 83.28 & 83.73 & 86.48 & 83.85 & 84.53 & 85.72 \\
   All predicate subimages & 82.79 & 83.33 & 86.37 & 83.30 & 84.22 &85.34 \\
  \bottomrule
  \end{tabular}
  }
 \label{tab:subimages}
\end{table}

\subsection{Baselines}
\noindent\textbf{CLIP}~\cite{clip} We use standard CLIP, where image embeddings are generated by CLIP’s vision encoder $F$; and text embeddings are generated by CLIP’s text encoder  $G$. The cosine similarity between them is computed to do matching.

\noindent\textbf{SLIP~\cite{slip}} 
We use the SLIP ViT-L-16. Similar to CLIP, the cosine similarity between the image embeddings and text embeddings is computed to do matching. 

\noindent\textbf{GLIP~\cite{GLIP}}
As GLIP has no global sentence and image embedding, we perform the following rule-based matching: 1) The image with more matched objects is predicted to be matching; 2) For images with the same set of objects, we compute the average confidence score of each object on both images. Larger score image is predicted. 

\noindent\textbf{BLIP2~\cite{blip2}}
We employed the official pretrained BLIP2. For the cosine similarity between image and text features, we adopted BLIP2's image-text contrastive learning match head as our BLIP2 baseline. Specifically, BLIP2 computes the cosine similarity score between each image embedding from each query output and the text embedding of the [CLS] token, selecting the highest similarity score as the ultimate outcome.

\subsection{Implementation Details}
The process begins by processing the original image with the dense caption module of GRiT~\cite{wu2022grit}, producing dense image captions based on object. The input text sentence is then parsed using the large language model (LLM), \texttt{gpt-3.5-turbo}, extracting entity words and organizing them into a subject-predicate-object format. We provide the prompt for parsing sentences for entities: \texttt{Analyze the objects in this sentence, the attributes of the objects and how each object is connected.} The prompt to match objects to text entities: \texttt{Find labels of the image that refer to this object from the sentence.} The alignment between dense image captions and entity words is realized using the same LLM, mapping entity words to their image counterparts based on captions.

For creating a predicate subimage, related object and subject subimages are combined. The original sentence and image, along with their respective parsed words and subimages, are fed into the CLIP text and vision encoders. Cosine similarity scores between each image and word embedding are computed and processed through a Softmax~\cite{softmax} layer, yielding three positive weights. The weighted sum of the subimage embeddings is then added to the original image's global embedding to obtain the final image embedding. The methodology remains similar for SLIP~\cite{slip} and BLIP2~\cite{blip2}, termed as ComSLIP and ComBLIP respectively. Notably, for BLIP2, we project the final image embedding to the sentence embedding dimension for the score computation.

\noindent\textbf{Evaluation Metrics~} 
We use Accuracy as the evaluation metric on the ComVG, SVO-Probes and VL-checklist datasets. For Winoground, we use three accuracy scores: text, image, and group score. The text score quantifies the proportion of both images correctly matched to their corresponding texts. The image score indicates the rate of both texts correctly matched to their corresponding images. Lastly, the group score signifies the accuracy of all texts and images matched correctly. We use Recall~\cite{recall} for Flickr30K and MSCOCO over the general image-text retrieval task.

\subsection{Main Results}
\noindent\textbf{Compositional Image and Text Matching}

\emph{Results on Winoground and VL-checklist} From Table~\ref{tab:winoground}, ComCLIP and ComSLIP consistently outperforms CLIP and SLIP respectively across both datasets, emphasizing their ability to grasp complex image-text relationships.
ComBLIP shows modest improvements, because BLIP2, pretrained on the Visual Genome dataset, already performs strongly.
Overall, it shows that our method's capability to be generalized to other stronger vision-language pretrained models.

\emph{Results on ComVG and SVO-Probes} In this subsection, we show the evaluation results on ComVG and SVO-Probes datasets in Table~\ref{tab:main}. Our ComCLIP can outperform zero-shot CLIP on both ComVG and SVO-Probes datasets. Separately reviewing the results, we see improvements in all negative types. This indicates that incorporating the information of subimages at inference time is helping CLIP attend to the semantic details of images and make fine-grained alignment. Apart from CLIP, we also validate the effectiveness of our method on SLIP~\cite{slip}, denoted by ComSLIP, with the results shown in Table~\ref{tab:main}. As presented, ours can beat SLIP on both the ComVG and SVO-Probes datasets, validating the effectiveness of our method on other CLIP-like models. In addition, we realize that our methods have lower performance improvement on the SVO-Probes dataset compared to ComVG on both CLIP and SLIP. This is because SVO-Probes contains sketchy data samples that we can not fully remove. We discuss some poor examples from SVO-Probes in the Appendix.

\emph{Comparison with GLIP} We compare our methods with GLIP in Table~\ref{tab:main}. Ours outperforms GLIP by a large margin on the compositional image-text matching task, further suggesting the effectiveness of our method compared with other region-based vision-language pretrained models.

\paragraph{General Image-Text Retrieval} 
Results on two image-text retrieval datasets are shown in Figure~\ref{fig:Flickr8K}. CLIP and ComCLIP both perform well in Recall@5, particularly in general image-text retrieval tasks like those in the Flickr30K, where compositionality comprehension is not crucial. ComCLIP outperforms CLIP in Recall@1 on both Flickr30K and MSCOCO, due to its focus on entities and their relations, steering CLIP away from decisions based on single nouns or spurious associations. Overall, these results suggest that our method is also competitive for general image-text retrieval tasks. 

\subsection{Ablations and Analysis}
\noindent\textbf{Ablation of Different Vision Encoders}
The results of using different vision encoders are shown in Table~\ref{tab:mergedResults}. ComCLIP demonstrates its effectiveness on various vision encoders and also yields notable improvements over OpenCLIP~\cite{openclip}, an open source implementation of CLIP. 

\begin{figure}[t]
    \centering
    \includegraphics[width=0.45\textwidth]{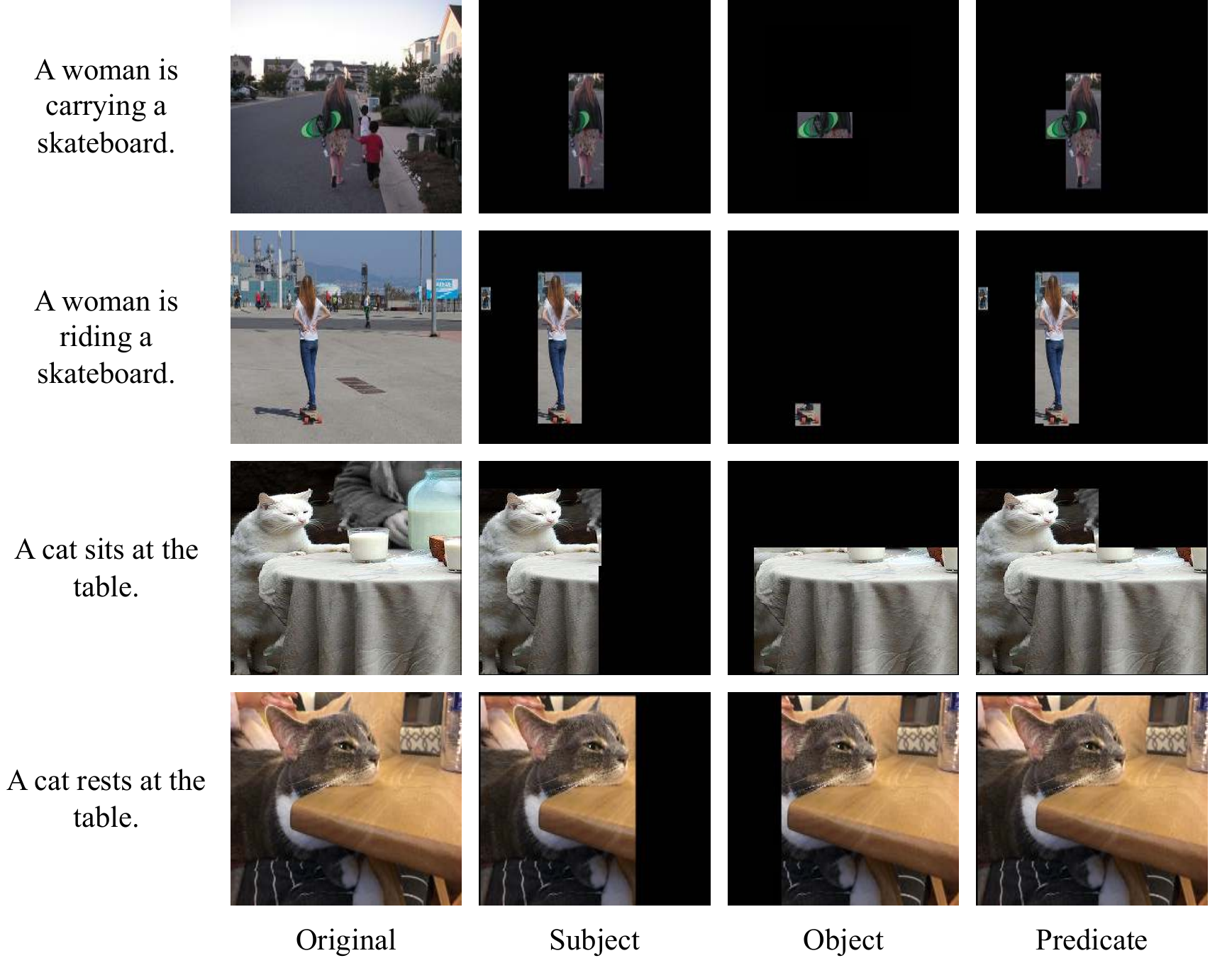}
    \caption{
    Examples of the generated subject, object, and predicate subimages. The first and third rows correspond to positive images and individual outputs of each IM for different entities. The second and fourth rows correspond to negative ones. \textbf{Top two rows}: examples from the ComVG dataset. (Woman, carrying, skateboard) is used as input (subject, predicate, object) to each IM.
    \textbf{Bottom two rows}: examples from the SVO-Probes dataset. (Cat, sits, table) is used as input to each IM. Note that for negative images, when IM could not accept the given (subject, predicate, object) and generate output subimages, the subimage is replaced with the original image for entity composition.
    }
    \label{fig:extracted+entity}
\end{figure}

\noindent\textbf{Ablation of Different Subimage Configurations}
Furthermore, in Table ~\ref{tab:subimages}, we show the efficacy of our method by comparing it against variations that employ either all black subimages or only one type of subimages. The results present that the amalgamation of subject, object, and predicate subimages achieved the highest accuracy across all vision encoders on both datasets. This implies that ComCLIP utilizes the specialized information conveyed by subimages to make accurate decisions.

\subsection{Qualitative Comparison}
 We illustrate the individual outputs of each IM for different entities in Figure~\ref{fig:extracted+entity}. In each row, we show from left to right: the original image $X$, subject image $\mathbf{S}$, object image $\mathbf{O}$, and predicate image $\mathbf{P}$.

\section{Conclusion} 
In this work, we observe that CLIP-like model could struggle in situations that require object, subject, and verb/predicate understanding when performing compositional image and text matching. Based on this observation, we propose a training-free method for compositional image and text matching from the causal view, mitigating the effect of spurious relations and improving compositional generalization. We also propose a new dataset to facilitate future research in this direction. 
Our method is plug-and-play and could be applied to other vision-language pretrained model. We hope that our simple yet effective training-free approach could boost the development of more interpretable and principled methods for the compositional image and text matching task.

\bibliography{anthology,custom}
\bibliographystyle{acl_natbib}

\clearpage

\appendix

This Appendix is organized as follows:
\begin{itemize}
    \item Section A contains a detailed process of creating subimages for image-text pairs;
    \item Section B compares the cost for running ComCLIP and CLIP;
    \item Section C contains a description of a causal graph for image-text matching;
    \item Section D contains additional implementation details of ComCLIP;
    \item Section E contains additional results on varied SVO-Probes data splits;
    \item Section F contains case studies from Flickr30K and MSCOCO;
    \item Section G contains an error analysis on SVO-Probes;
    \item Section H contains experiments of ComCLIP's performance with different language parsers;
    \item Section I contains an evaluation of GRiT's robustness as our counterfactual subimage generator;
    \item Section J contains ablation experiments with additional counterfactual subimage generators;
    \item Section K contains ablation experiments on all except on one type subimages;
    \item Section L contains experiments presenting ComCLIP's superiority to simple entity-image matching;
    \item Section M contains data examples from MSCOCO;
    \item Section N contains data examples from Winoground, ComVG and SVO-Probes;
    \item Section O contains comparison between ComCLIP and finetuned ComCLIP;
    \item Section P contains experiments of applying our methods to Instance-level Image-text Matching Baselines;
    \item Section Q summarizes the detailed algorithm of ComCLIP.
\end{itemize}

\begin{figure*}[t]
    \centering
    \includegraphics[width=0.9\textwidth]{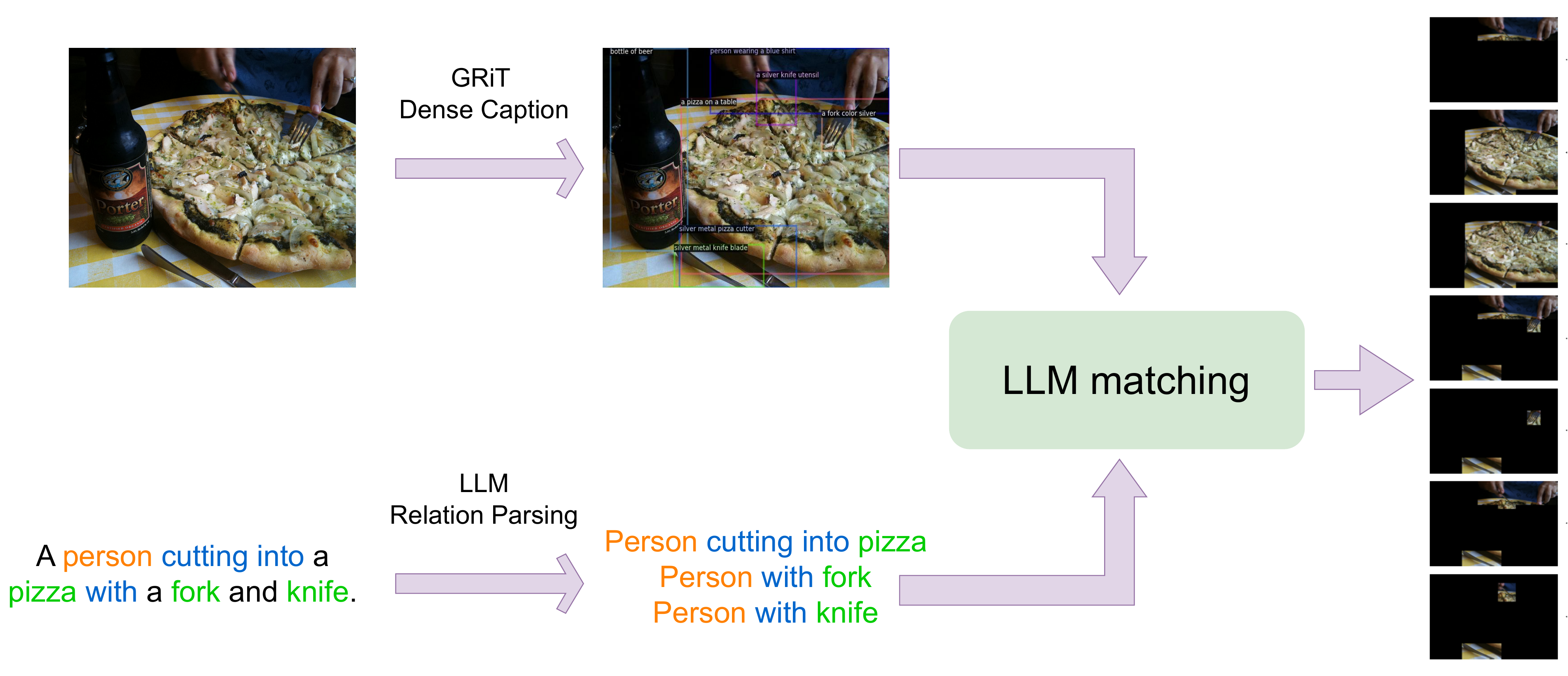}
    \caption{Design of the counterfactual subimage generation process. LLM matches the dense captions generated by GRiT from image to parsed subjects, objects, predicates from text.}
    \label{fig:grit_llm}
\end{figure*}

\begin{figure}[t]
    \centering
    \includegraphics[width=0.2\textwidth]{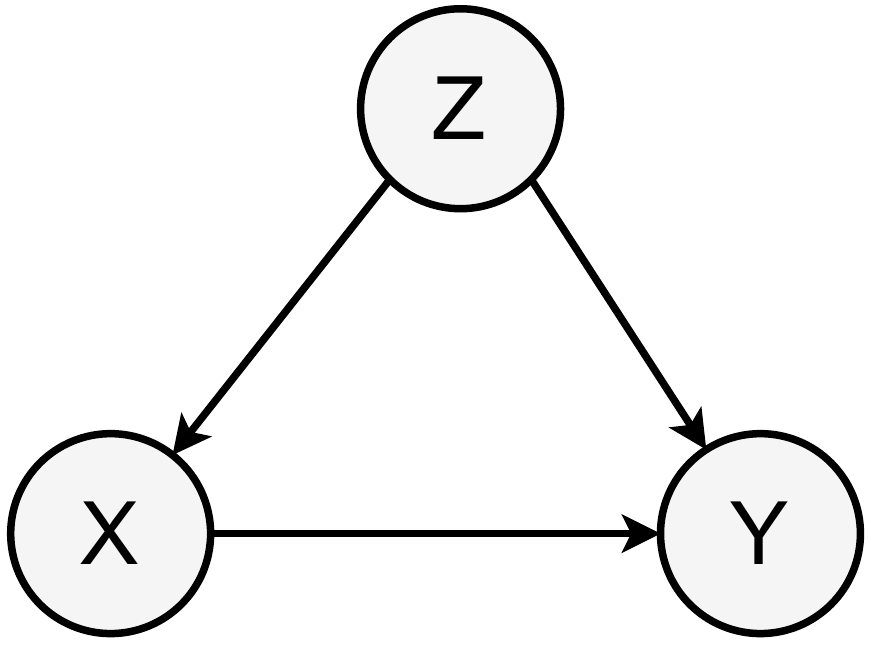}
    \caption{The causal graphs in the context of compositional image-text matching.
    }
    \label{fig:causal}
\end{figure}

\section{Counterfactual Subimage Generation}
Figure~\ref{fig:grit_llm} presents a visual guide to our subimage creation process for image-text pairs. For instance, GRiT analyzes the image, generating detailed captions for objects such as \textcolor[RGB]{0, 128, 0}{pizza}, \textcolor{orange}{person}, \textcolor[RGB]{0, 128, 0}{fork}, and \textcolor[RGB]{0, 128, 0}{knife}, along with their spatial references. Next, LLM extracts relation triplets from sentences, like \textit{\textcolor{orange}{person} \textcolor{blue}{cutting into} \textcolor[RGB]{0, 128, 0}{pizza}} and \textit{\textcolor{orange}{person} \textcolor{blue}{with} \textcolor[RGB]{0, 128, 0}{a fork}}. Utilizing LLM again, we identify all captions that could pertain to an object. To illustrate, for creating the \textcolor[RGB]{0, 128, 0}{pizza} subimage, LLM recognizes that the dense caption \textit{a pizza on a table} refers to \textcolor[RGB]{0, 128, 0}{pizza}, so we use the corresponding image section of this caption. For generating the predicate \textcolor{blue}{cutting into} subimage, we merely overlap the subimages of \textcolor{orange}{person} and \textcolor[RGB]{0, 128, 0}{pizza}, the subject and object of \textcolor{blue}{cutting into} respectively.

\section{Inference Cost}
\label{inferencecost}
This section offers a comparative analysis of the inference time for processing a single image-text pair using ComCLIP and the standard CLIP model. The evaluation, conducted over 10 trials with four V100 GPUs, incorporated pre-extracted subimages and entity words to optimize the process. The results indicate that the average inference time for the CLIP model is 0.24$\pm$0.01 seconds, while for our ComCLIP model, it is marginally higher at 0.25$\pm$0.03 seconds using the ViT-B/32 architecture. This minor increase is particularly noteworthy as it falls within the same order of magnitude, underscoring the efficiency of ComCLIP in maintaining comparable processing speeds.

Furthermore, the GPU memory consumption during inference was also assessed. The CLIP model utilized 2047$\pm$44 MB, and ComCLIP required slightly more at 2086$\pm$98 MB. This modest increment in memory usage is offset by the enhanced capabilities of ComCLIP, affirming its practicality for deployment in similar computational settings. Thus, ComCLIP stands out as an efficient solution, offering advanced functionalities with only a nominal increase in resource requirements.


\section{Causal Graph in the Context of Image-text Matching}
We show the causal graph in the context of our image-text matching task in Figure~\ref{fig:causal}. $X$ are high-dimensional observations (i.e., images), and $Y$ are corresponding text prompts. $X$ can be described by lower-dimensional, semantically meaningful factors of variation $Z$ (e.g., objects, subjects, or action relations between objects and subjects (i.e., predicates in the image)). 

\section{Implementation Details}
We introduce the implementation details of ComCLIP in this section. Our pipeline is training-free, so there are no parameters involved in ComCLIP. In the main paper, we use the CLIP model with a ViT-B-32 vision encoder for the results in Table~\ref{tab:winoground}, and a ViT-L-14 vision encoder for the results in Table~\ref{tab:main}. The masks for subjects/objects/predicates are generated using GRiT~\cite{wu2022grit} with the dense caption version, which is pre-trained for 200 epochs.

\section{Experimental Results on SVO-Probes over Different Splits}
\begin{table*}[t]
  \centering
  \caption{Comparison of ComCLIP with CLIP under three different splits on the SVO-Probes dataset.   }
  \begin{tabular}{lcccccc}
    \toprule
    & \multicolumn{2}{c}{Seed 42} & \multicolumn{2}{c}{Seed 11} & \multicolumn{2}{c}{Seed 2}\\
     \cmidrule(lr){2-3}\cmidrule(lr){4-5}\cmidrule(lr){6-7}
    Vision Encoder & {CLIP}& {ComCLIP} & {CLIP}& {ComCLIP}& {CLIP}& {ComCLIP}\\
  \midrule
   ResNet-50 &82.77 &83.87&82.06&83.10&82.87&83.97\\
   ViT-B-32 &84.13&85.47&84.47&84.83&84.17&84.67\\
   ViT-L-14 &85.53 &86.63 &84.76 &86.10&85.27&86.33\\
  \bottomrule
  \end{tabular}
  \label{seed_results}
  \end{table*}
  
\begin{table*}[t]
  \centering
  \caption{Effectiveness of our method using SLIP under three different splits on the SVO-Probes dataset.}
  \begin{tabular}{lcccccc}
    \toprule
    & \multicolumn{2}{c}{Seed 42} & \multicolumn{2}{c}{Seed 11} & \multicolumn{2}{c}{Seed 2} \\
     \cmidrule(lr){2-3}\cmidrule(lr){4-5}\cmidrule(lr){6-7}
    Vision Encoder & {SLIP}& {ComSLIP} & {SLIP}& {ComSLIP} & {SLIP}& {ComSLIP} \\
  \midrule
   SLIP (ViT-B-32)& 77.70&77.90 &{79.10}&{79.75}&81.00&80.15\\
   SLIP (ViT-L-14)& 78.90&79.70&{79.70}&{80.15}&80.10&81.30\\
     \bottomrule
  \end{tabular}
 \label{tab:other}
\end{table*}
In this section, we show additional results using three different data splits on SVO-Probes. We use random seeds $42, 11, 2$ to re-split the dataset, with the results of CLIP vs. ComCLIP shown in Table~\ref{seed_results} and the results of other CLIP-based models shown in Table~\ref{tab:other}.


\begin{algorithm}[t]
\caption{Training-Free Compositional Image and Text Matching with ComCLIP.}
\label{algo:algo}
\begin{algorithmic}[1]

\REQUIRE ~~\\
\textbf{Input:} image $X$, text prompt $Y$, vision encoder $F(\cdot)$, text encoder $G(\cdot)$, independent mechanisms $f_{\text {object }}(\cdot), f_{\text {subject}}(\cdot), f_{\text {predicate}}(\cdot)$.\\
\textbf{Output:} Matching score $O$.\\

\STATE Generate counterfactual subimages \\
$\mathbf{O}, \mathbf{S}, \mathbf{P} \!\!\leftarrow \!\! f_{\text {object }}(X), f_{\text {subject}}(X), f_{\text {predicate}}(X)$; \\

\STATE Extract feature embeddings \\
$F(\mathbf{O}), F(\mathbf{S}), F(\mathbf{B}) \leftarrow \mathbf{O}, \mathbf{S}, \mathbf{P}$;\\
\STATE Extract (subject, object, predicate) words \\ $Y_s, Y_o, Y_p \leftarrow Y$; \\
\STATE Compute the concept-level similarity $S_1, S_2, S_3 \leftarrow G(Y_s), G(Y_o), G(Y_p), F(\mathbf{O}), F(\mathbf{S}), F(\mathbf{P})$; \hfill\COMMENT{Eq. (3)} \\
\STATE Extract sentence embeddings \\
$G(Y) \leftarrow Y$;   \\
\STATE Compose visual features $V \leftarrow S_1,  S_2, S_3$, \\ $f_{\text {object }}(\cdot), f_{\text {subject}}(\cdot), f_{\text {predicate}}(\cdot), F(\cdot), X; $ \hfill\COMMENT{Eq. (4)} \\
\STATE Compute the matching score $O \leftarrow Y, V $  \\
\end{algorithmic}
\end{algorithm}

\section{Case Study: Generalized Scenario with Multiple SVO}
In this section, we present the case study where the text contains multiple SVOs on Flickr30K and MSCOCO. 

\subsection{Cases Study on Flickr30K}
\begin{figure*}[t]
    \centering
    \includegraphics[width=\textwidth]{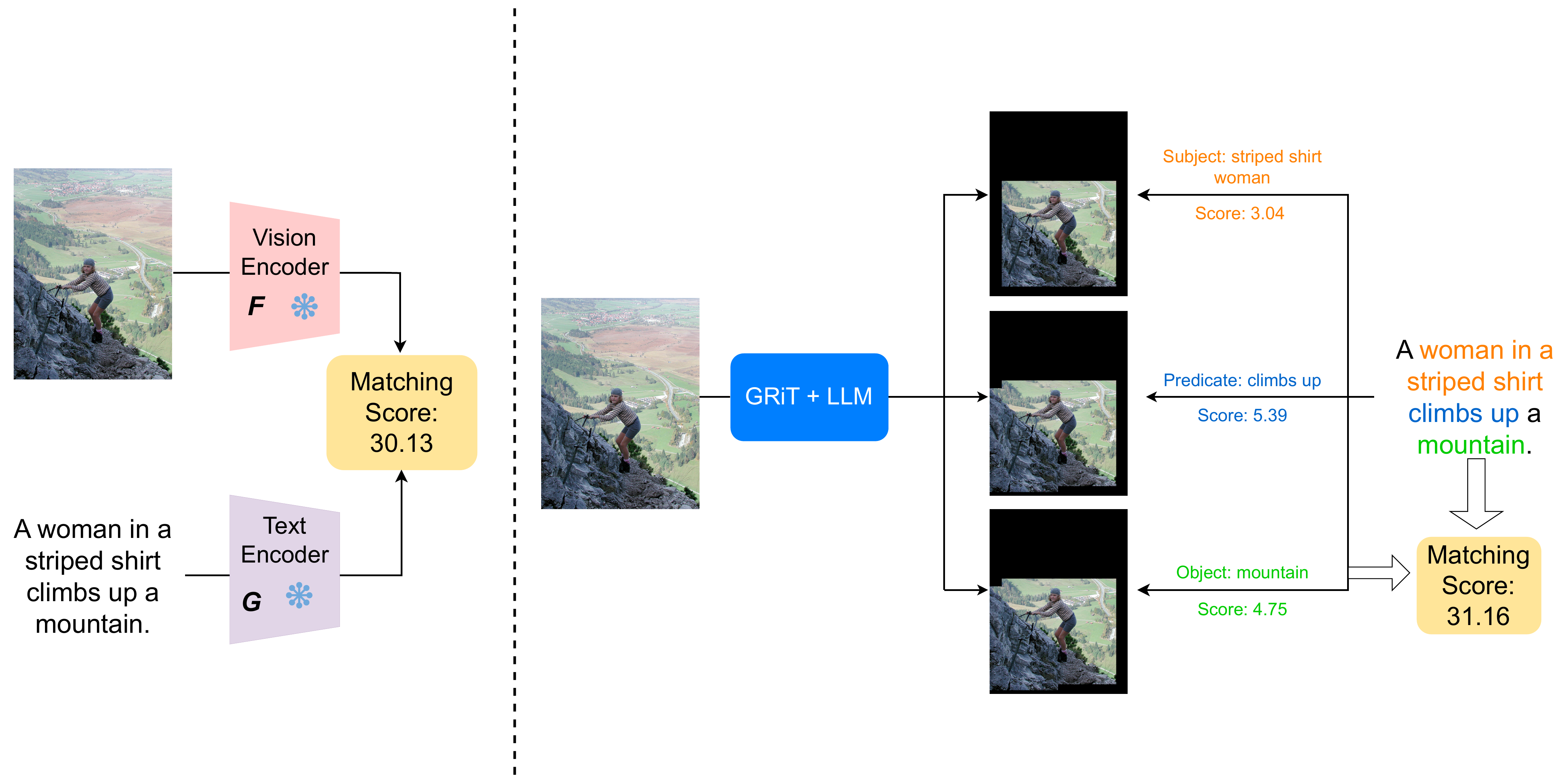}
    \caption{The comparison of CLIP (left) and ComCLIP (right) over the case where single subjects/objects/predicates are involved. Image and text examples are from Flickr30K.
    }
    \label{case2}
\end{figure*}

In Figure~\ref{case2}, we first show the case where single SVO are involved.

\begin{figure*}[t]
    \centering
    \includegraphics[width=\textwidth]{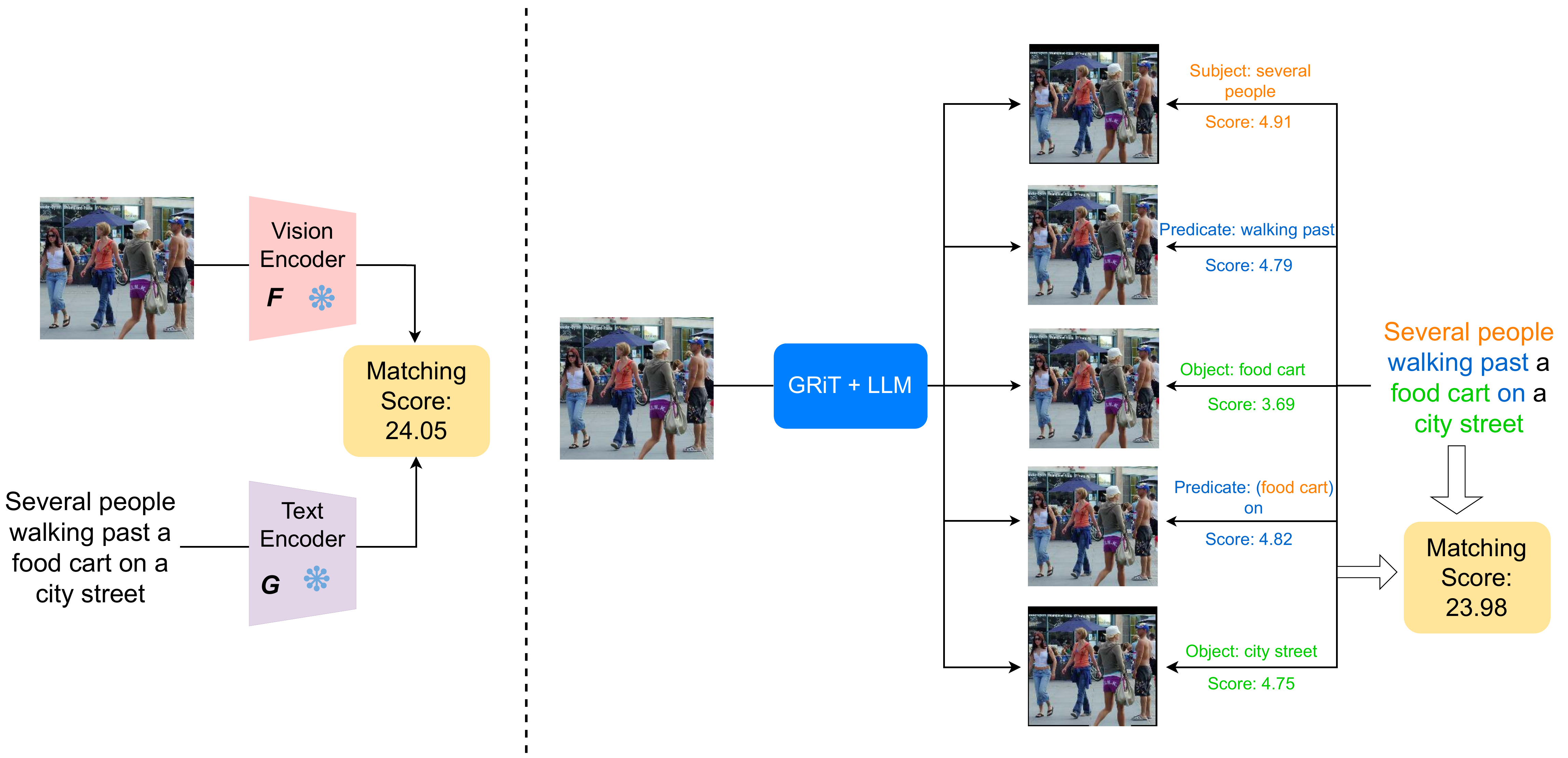}
    \caption{The comparison of CLIP (left) and ComCLIP (right) over the case where multiple subjects/objects/predicates are involved (this is a negative example from Flickr30K). 
    }
    \label{case_study3}
\end{figure*}

In Figure~\ref{case_study1} and Figure~\ref{case_study3}, we show the case where multiple SVOs are involved. Specifically, in this provided case, multiple objects (\textcolor[RGB]{0, 128, 0}{Food cart, City street}) and subjects (\textcolor{orange}{Several People, Food cart}) are involved.  Figure~\ref{case_study1} is a positive example, and Figure~\ref{case_study3} is a negative example. As can be seen, ComCLIP can utilize multiple subjects/objects/predicates in the input texts to do the matching. The \textcolor[RGB]{0, 128, 0}{food cart} object dominates the decision process and helps ComCLIP make the correct match.

\subsection{Case Study on MSCOCO}
\begin{figure*}[t]
    \centering
    \includegraphics[width=\textwidth]{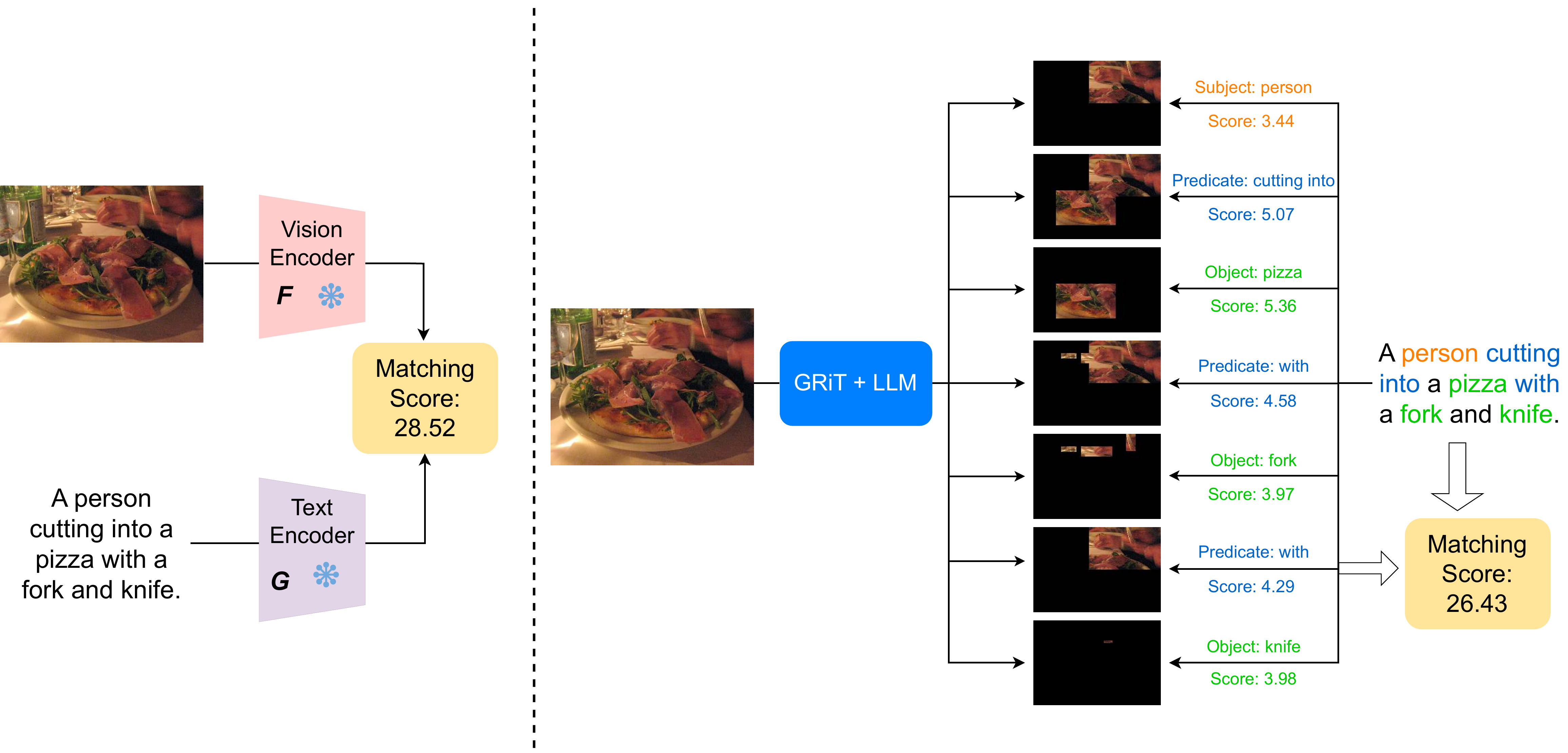}
    \caption{The comparison of CLIP (left) and ComCLIP (right) over the case where multiple subjects/objects/predicates are involved (this is a negative example from MSCOCO).
    }
    \label{fig:mscoco_wrong}
\end{figure*}

\begin{figure*}[t]
    \centering
    \includegraphics[width=\textwidth]{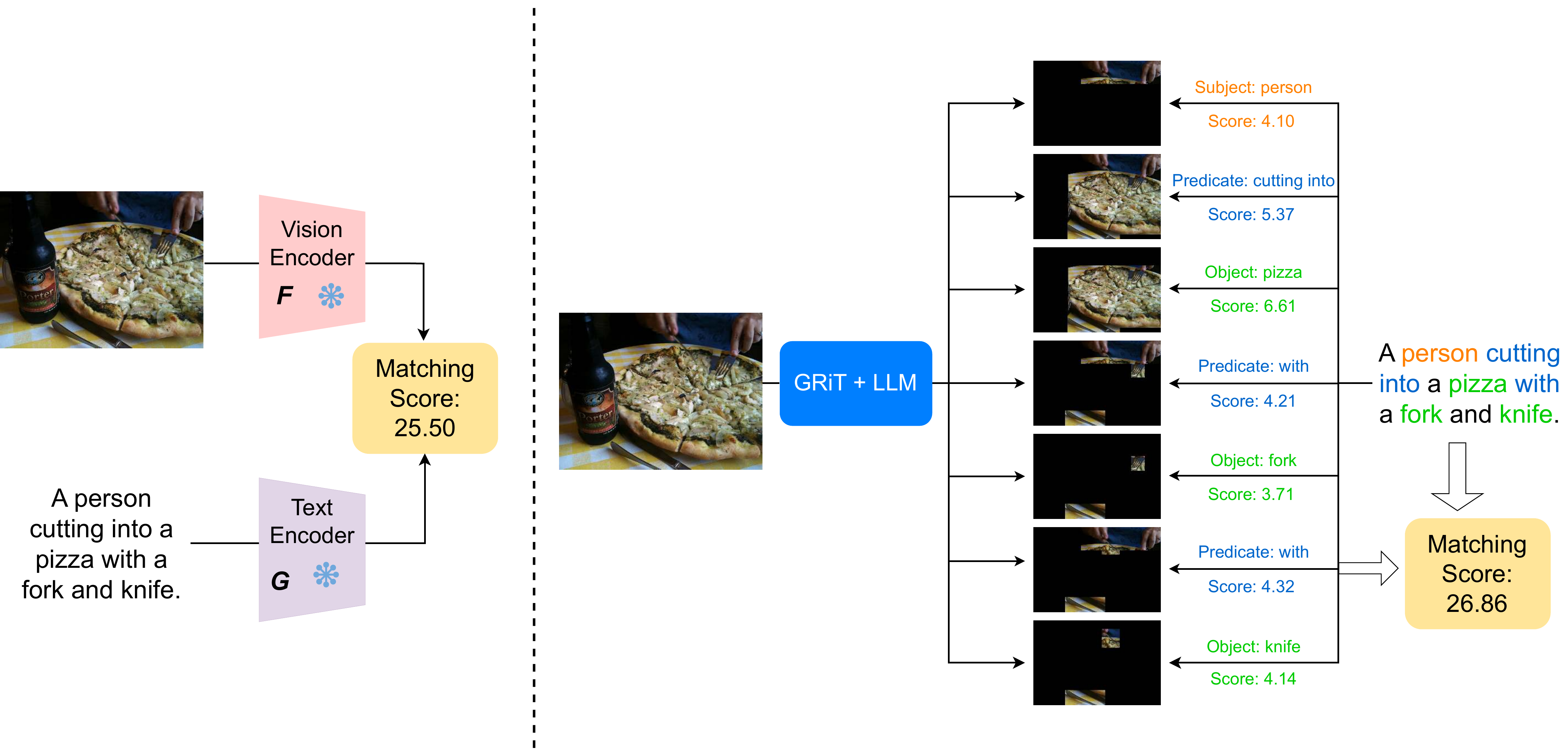}
    \caption{The comparison of CLIP (left) and ComCLIP (right) over the case where multiple subjects/objects/predicates are involved (this is a positive example from MSCOCO).}
    \label{fig:mscoco_correct}
\end{figure*}

In Figure~\ref{fig:mscoco_wrong} and \ref{fig:mscoco_correct}, we provide a breakdown of how ComCLIP makes the correct decision when multiple SVOs are involved on MSCOCO. Both the negative image from Figure~\ref{fig:mscoco_wrong} and the positive image from Figure~\ref{fig:mscoco_correct} are closely aligned with the text, featuring prominent visual entities such as a person and a pizza. ComCLIP integrates various subjects, objects, and predicates, effectively distinguishing the correct image match from a pair of visually analogous images.

\begin{figure}[htbp]
    \centering
    \includegraphics[width=0.5\textwidth]{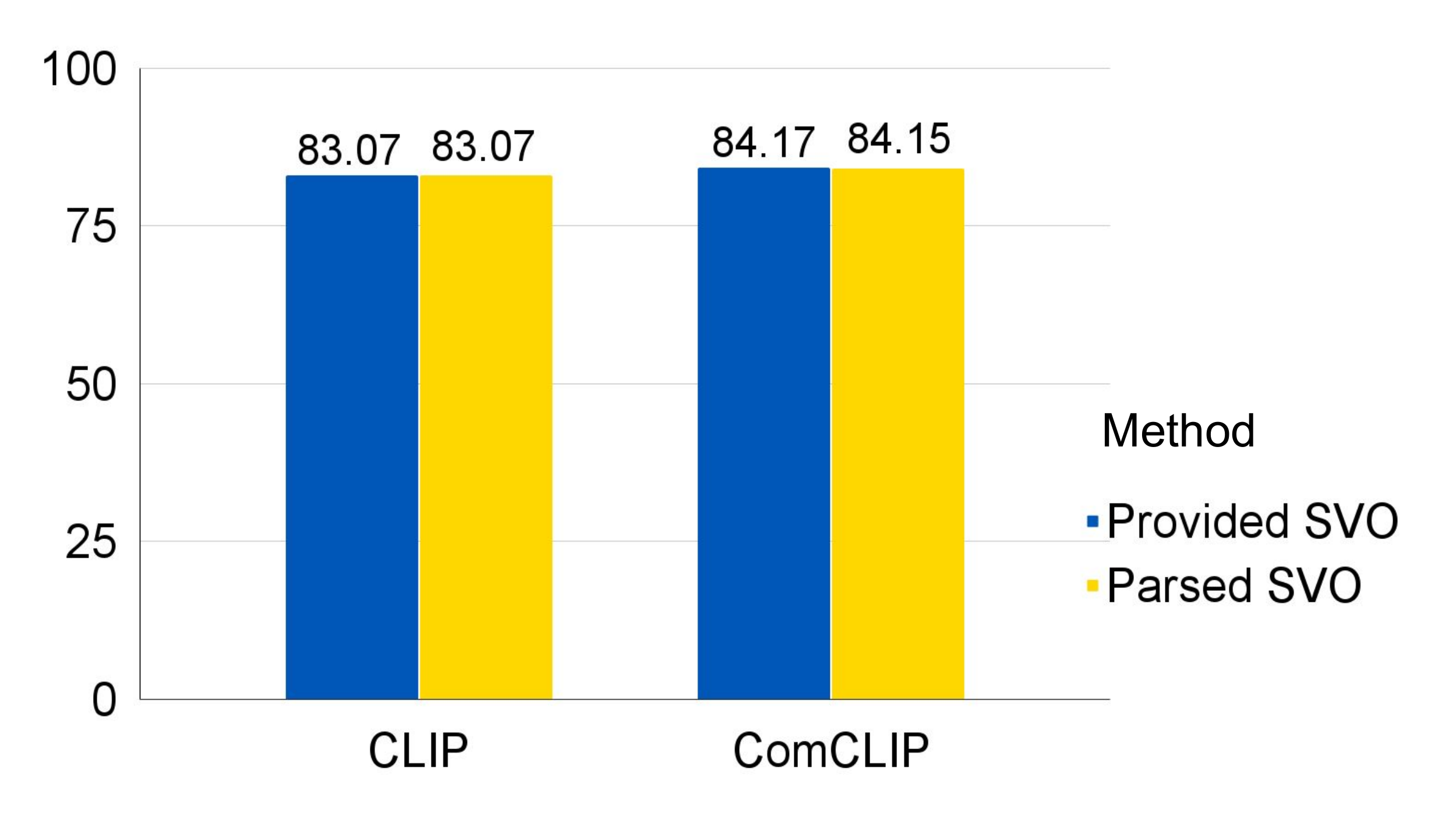}
    \caption{Comparison of accuracy (\%) on SVO-Probes using parsed and ground-truth SVO triplets.
    }
    \label{fig:parsing}
\end{figure}

 \begin{table*}[t]
  \centering
   \caption{Compositional Visual Genome subset accuracy  (\%) with masks generated by Lang-Seg and CLIPSeg.}
    \begin{tabular}{lccc}  
    \toprule
    & {ResNet-50} &{ViT-B-32} &{ViT-L-14} \\
  \midrule
   CLIP &79.38 & 79,94 &83.70 \\
   ComCLIP (Lang-Seg mask) & 82.41 & \textbf{83.15}  & \textbf{86.05} \\
    ComCLIP (Lang-Seg mask with blur) & 83.27 & 83.09  & 85.31 \\
    ComCLIP (CLIPSeg mask) & \textbf{83.27} &  82.22 & 85.18 \\
    ComCLIP (CLIPSeg mask with blur) & 82.78 & 82.90  & 85.31 \\
  \bottomrule
   \end{tabular}
  \label{clipseg}
  \end{table*}

\section{Error Analysis
}
\label{bad_example}
\begin{figure*}[t]
    \centering
    \includegraphics[width=0.9\textwidth]{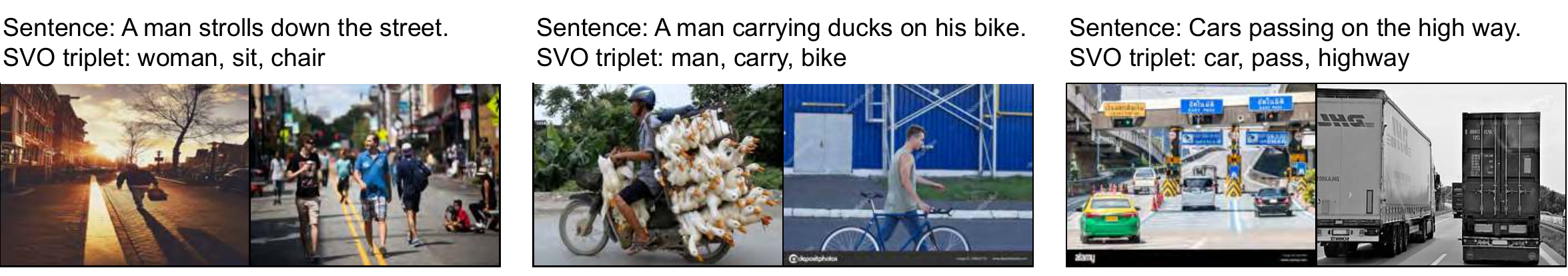}
    \caption{Selected bad quality examples from the SVO-Probes dataset.}
    \label{fig:svo}
\end{figure*}

As shown in the main paper, we get higher improvements using ComCLIP on Compositional Visual Genome compared with SVO-Probes. This is mainly because our collected Compositional Visual Genome is cleaner and the SVO-Probes dataset tends to be noisy. Herein, we give a case study covering three major error-inducing issues found in SVO-Probes, as depicted in Figure~\ref{fig:svo}: instances where the negative image aligns with the input sentence, object mismatches, and the presence of watermarks in images.

\begin{figure*}[t]
    \centering
    \includegraphics[width=\textwidth]{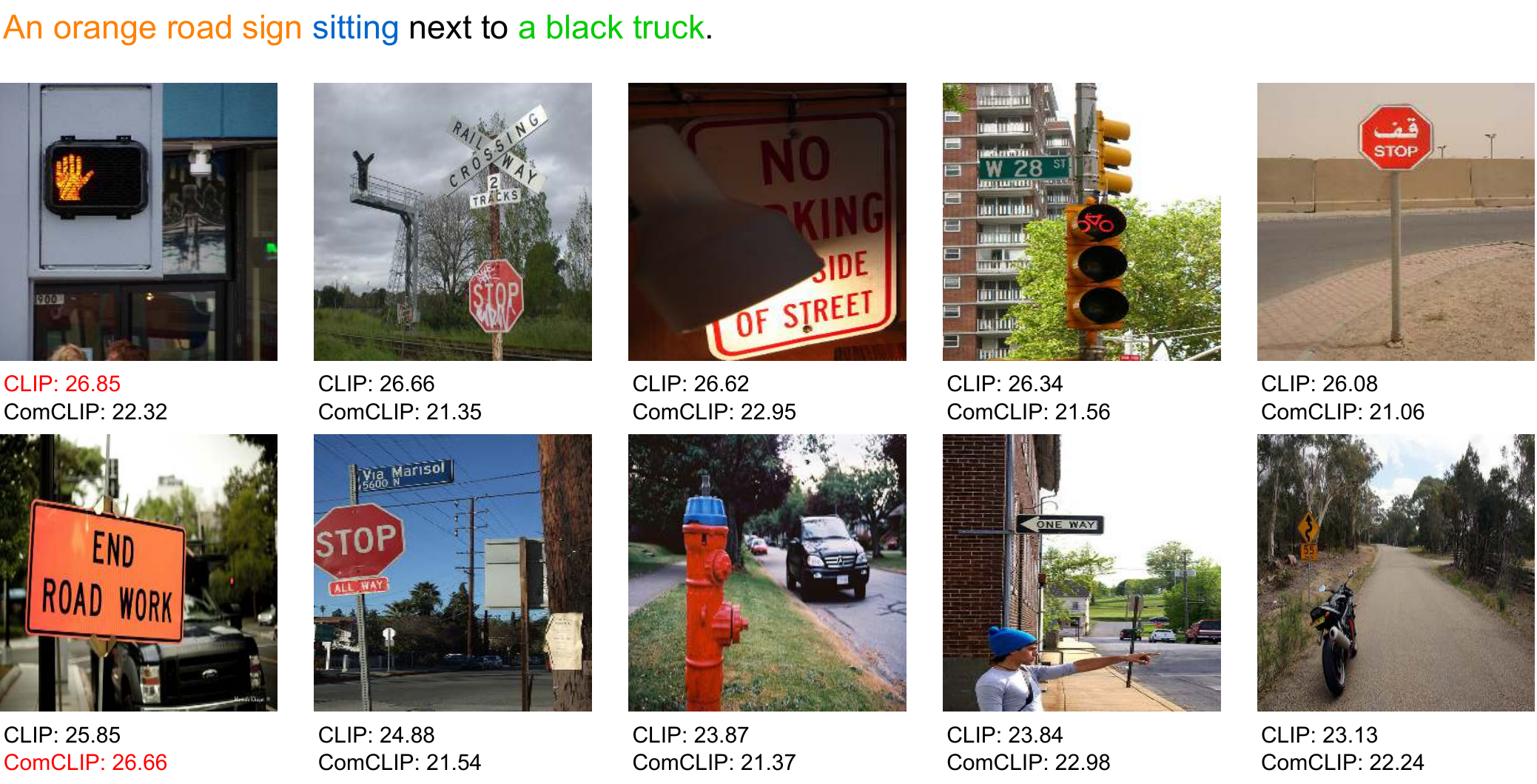}
    \caption{Example from MSCOCO dataset.}
    \label{fig:mscoco_ten_image}
\end{figure*}
\begin{figure*}[t]
    \centering
    \includegraphics[width=0.95\textwidth]{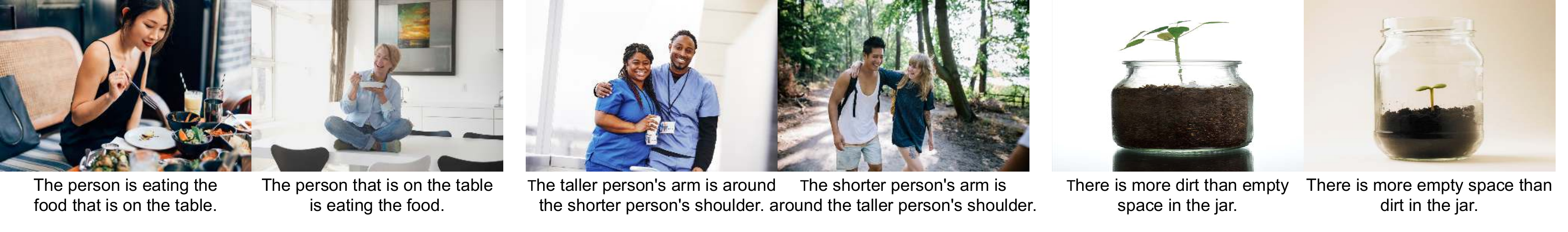}
    \caption{Examples from Winoground dataset.}
    \label{fig:wino}
\end{figure*}
\begin{figure*}[t]
    \centering
    \includegraphics[width=0.95\textwidth]{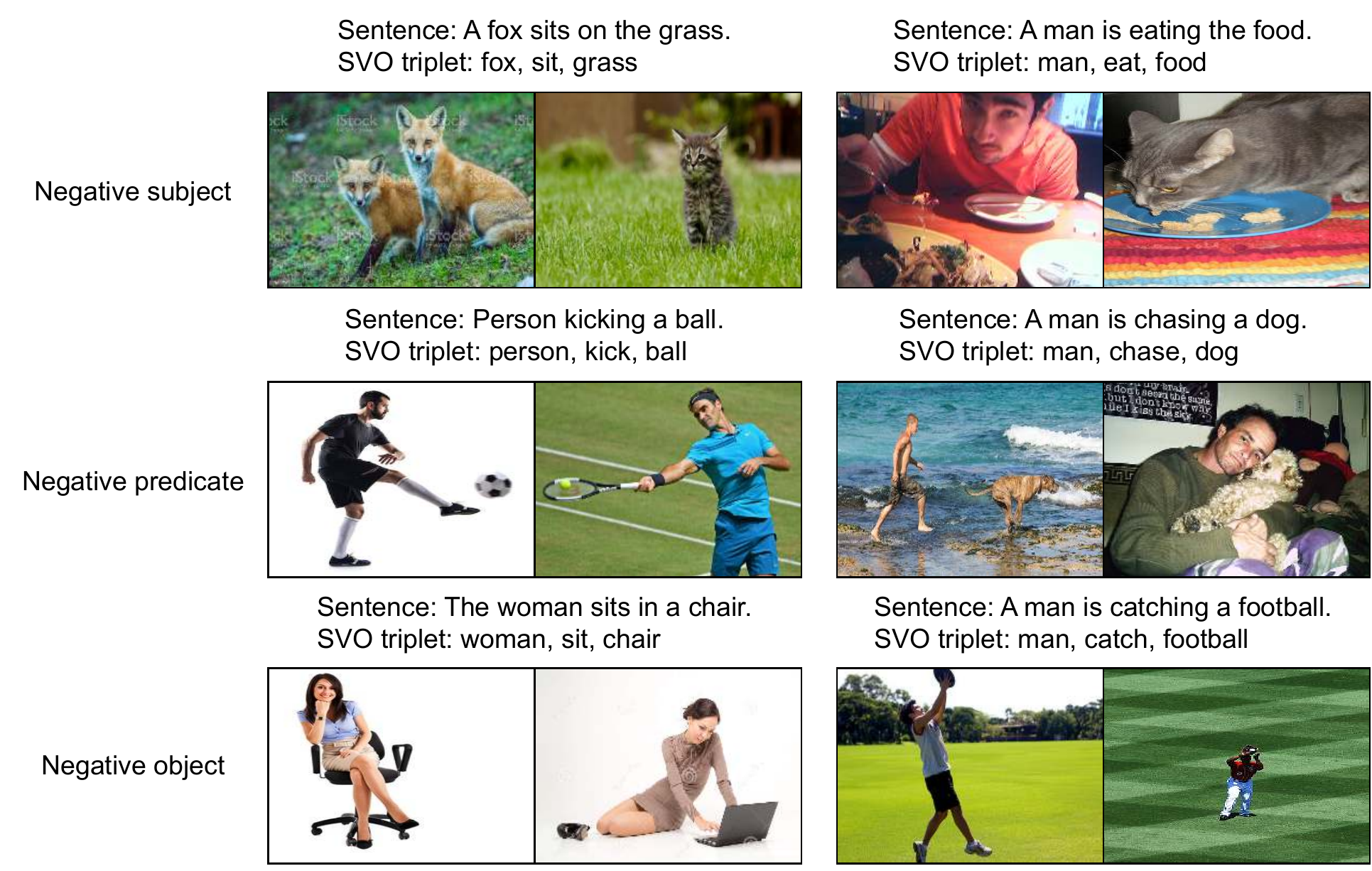}
    \caption{Examples from the SVO-Probes dataset (left two) and Compositional Visual Genome dataset (right two). There are three negative types for a given triplet: a subject-negative, predicate-negative, or object-negative where respectively, the subject, predicate, or object in the triplet is replaced by a different word. Within each image pair, the positive image on the left represents the positive triplet, while the negative image on the right corresponds to the negative triplet.}
    \label{fig:svo_vg}
\end{figure*}

\section{Extracted Entities}
\paragraph{Use Language Parser to Extract SVO}
The performance of ComCLIP is also dependent on the quality of the subject,
object, and predicate entity provided. To study the effect of extracted entities, we analyze our methods on SVO-Probes since it has more complex sentence structures. Apart from the LLM approach shown in the main paper, we remove stop words from the sentence using NLTK~\cite{nltk} and then use a Subject Verb Object extractor developed based on~\cite{spacy2} to extract the subject, predicate, and object from the original sentence. Figure~\ref{fig:parsing} shows that our parsed entities have almost the same performance as that using the ground truth subjects, predicates, and objects.

\section{Robustness of Counterfactual Subimage Generator}
To show the robustness of using GRiT~\cite{wu2022grit} to generate counterfactual subimages, we quote the results and conclusions from~\cite{wu2022grit}.
According to~\cite{wu2022grit}, GRiT is comparable to the closed-set object detector with a 0.8 AP gap. This result demonstrates GRiT’s open-set framework can serve as a new promising formulation for
object detection. GRiT also performs comparably with the
state-of-the-art closed-set object detectors. This once again
demonstrates GRiT can serve as the subimage generator in our pipeline.

\section{Additional Ablations on Counterfactual Subimage Generation} In this section, we show that ComCLIP is \textbf{robust} to the choice of counterfactual subimage generator. We use segmentation models, Lang-Seg and CLIPSeg~\cite{clipseg} with the \texttt{clipseg-rd64-refined} version, to create segmentation masks and generate subimages. Specifically, given the input  (subject, object, predicate) triplet, we model the object mechanism $f_{\text {object }}$ using a binary mask generated by Lang-Seg and CLIPSeg, which are both CLIP-based language-guided segmentation models.  Given the segmentation results, the object part will be set to $1$ while the remainder of the image is $0$.
In a manner similar to the object mechanism, the subject mechanism $f_{\text {subject}}$ is achieved by setting the background to $0$ while the subject region is set to $1$.
The predicate mechanism $f_{\text {predicate}}$ is implemented by combining the binary mask generated by $f_{\text {object }}$ and $f_{\text {subject}}$ together: the object and subject regions will be $1$ while the remaining regions will be $0$. 

We test the masks on a randomly selected 30\% subset of Compositional Visual Genome. The results in Table~\ref{clipseg} indicate that ComCLIP continues to outperform CLIP across all vision encoders when the masks are generated by CLIPSeg. To further test its robustness, we add noise by applying Gaussian image blur to the backgrounds of generated subimages rather than using pure black backgrounds. Despite the blurring, ComCLIP using either Lang-Seg or CLIPSeg masks still performs better than CLIP and achieves similar performance to ComCLIP without blur as shown in Table ~\ref{clipseg}. Thus, ComCLIP is shown to be resilient to the precision of generated subimages.

 \section{Additional Ablations on All Except One Type Subimages} We test 3 combinations of subimages on a balanced randomly sampled 3000 subset of ComVG, presented in Table~\ref{tab:sub_obj_ablation}. As can be observed, ComCLIP outperforms all 3 scenarios in which all but one subimage are utilized, confirming that ComCLIP effectively leverages the composite information for reasoning.\\
\begin{table*}[t]
  \caption{Results of different subimg configurations (ComVG)}
  \centering
  \begin{tabular}{lcccc}
    \toprule
    Vision Encoder& All Sub \& Obj & All Sub \& Pred & All Obj \& Pred & ComCLIP $\spadesuit$ \\
    \midrule 
    ResNet-50 & 81.03 \scriptsize $\mathcolor{darkgreen}{(\textbf{-0.04})}$ & 80.47 \scriptsize $\mathcolor{darkgreen}{(\textbf{-0.60})}$ & 79.73 \scriptsize $\mathcolor{darkgreen}{(\textbf{-1.34})}$ & \textbf{81.07} \\
    ViT-B-32 & 80.70 \scriptsize $\mathcolor{darkgreen}{(\textbf{-0.10})}$ & 79.33 \scriptsize $\mathcolor{darkgreen}{(\textbf{-1.47})}$ & 80.60 \scriptsize $\mathcolor{darkgreen}{(\textbf{-0.20})}$ & \textbf{80.80} \\
    ViT-L-14 & 84.37 \scriptsize $\mathcolor{darkgreen}{(\textbf{-0.06})}$ & 83.33 \scriptsize $\mathcolor{darkgreen}{(\textbf{-1.10})}$ & 83.73 \scriptsize $\mathcolor{darkgreen}{(\textbf{-0.70})}$ & \textbf{84.43} \\
    \bottomrule
  \end{tabular}
 \label{tab:sub_obj_ablation}
\end{table*}

\begin{table*}[t]
  \caption{Fine-grained similarity w/ parsed words (ComVG)}
  \centering
  \begin{tabular}{lccccc}
    \toprule
    {Vision Encoder}& {Subject Entity} &{Predicate Entity} &{Object Entity} & {All Entity} & {ComCLIP $\spadesuit$}\\
       \midrule 
   ResNet-50 & 61.37 \scriptsize $\mathcolor{darkgreen}{(\textbf{-22.36})}$ 
   & 52.98 \scriptsize $\mathcolor{darkgreen}{(\textbf{-30.75})}$& 68.76 \scriptsize $\mathcolor{darkgreen}{(\textbf{-14.97})}$&81.17 \scriptsize $\mathcolor{darkgreen}{(\textbf{-2.56})}$&\textbf{83.73}\\
   Vit-B-32 & 60.57 \scriptsize $\mathcolor{darkgreen}{(\textbf{-24.18})}$  & 53.35 \scriptsize $\mathcolor{darkgreen}{(\textbf{-31.40})}$&69.91 \scriptsize $\mathcolor{darkgreen}{(\textbf{-14.84})}$&81.44 \scriptsize $\mathcolor{darkgreen}{(\textbf{-3.31})}$& \textbf{84.75}\\
   Vit-L-14 & 62.17 \scriptsize $\mathcolor{darkgreen}{(\textbf{-25.23})}$& 54.98  \scriptsize $\mathcolor{darkgreen}{(\textbf{-32.42})}$& 70.52 \scriptsize $\mathcolor{darkgreen}{(\textbf{-16.88})}$& 84.85 \scriptsize $\mathcolor{darkgreen}{(\textbf{-2.55})}$&\textbf{87.40}\\
  \bottomrule
  \end{tabular}
 \label{tab:single_entity}
\end{table*}

 \section{Additional Ablations on Comparisons with Fine-grained Similarity Matching Methods} 
In our additional analysis, detailed in Table ~\ref{tab:single_entity}, we explore the impact of matching individual parsed entity words with images, as opposed to full sentences, employing the CLIP architecture as our foundation. The results demonstrate that ComCLIP markedly surpasses the performance of three baseline models on four entity scenarios, which are based solely on the similarity between a single entity word and an image. This highlights the superior efficacy of ComCLIP.

\section{Data Examples from MSCOCO}
In this section, we provide an example from the MSCOCO dataset that we constructed, as shown in Figure~\ref{fig:mscoco_ten_image}. The MSCOCO dataset typically incorporates adjectives to enhance query sentences, which CLIP tends to overlook. For instance, in the provided example, the \textcolor{orange}{orange road sign} helps ComCLIP successfully identify the accurate image as the best match, while CLIP does not rank it among the top 5 matches.

\section{Data Examples from Winoground, ComVG and SVO-Probes}
\label{e1}
In this section, we show examples from Winoground in Figure~\ref{fig:wino}. Winoground presents a challenging task, requiring precise match of two image-text pairs to successfully earn a group score. We also show the ComVG dataset constructed by us and the SVO-Probes in Figure~\ref{fig:svo_vg}. As can be seen, they are formatted similarly: Negative Types --- Sentence --- SVO Triplet --- Positive Image --- Negative Image. Visual Genome is licensed under a Creative Commons Attribution 4.0 International License.
Compositional Visual Genome dataset is compatible with the original access conditions of Visual Genome. 

\section{Compared with Finetuned ComCLIP}

In addition to the original ComCLIP model, we explored the effects of finetuning ComCLIP using the MSCOCO dataset, subsequently evaluating its performance on the ComVG dataset and SVO-Probes dataset. This process involved a approach to training example construction: for each query text, we utilized the CLIP model to identify the most challenging negative image from the MSCOCO training set, based on the highest similarity score. This method aimed to enhance the model's ability to discern subtle distinctions between closely related visual-textual pairs. The resulted finetuned CLIP is still
evaluated in a zero-shot fashion on the target evaluation dataset, i.e., how well does finetuned ComCLIP trained on MSCOCO transfer to target datasets. The results, as outlined in Table~\ref{tab:finetunecomclip}, demonstrate notable improvements in the finetuned ComCLIP's performance compared to both the standard CLIP and the unfinetuned ComCLIP models.

The finetuned ComCLIP model exhibited significant gains across all categories on both the ComVG and SVO-Probes datasets. Particularly, the average accuracy on the ComVG dataset increased from 84.63\% for ComCLIP to 86.98\% for the finetuned version, underscoring the effectiveness of finetuning in enhancing model performance. Similarly, on the SVO-Probes, there was an increase from 86.41\% to 87.99\%. These improvements are most prominent in the `Object' category of the ComVG dataset, where the finetuned ComCLIP achieved a 97.83\% accuracy, indicating a substantial enhancement over the original model's performance.

These results suggest that finetuning on a dataset with diverse visual and textual representations, such as MSCOCO, significantly improves the model's capability to generalize and transfer learned features to different, yet related, datasets. The enhancements in accuracy, particularly in the `Object' recognition category, could be attributed to the comprehensive and varied nature of objects represented in the MSCOCO dataset, which may have provided a more robust learning experience for the model.

This analysis indicates that while the original ComCLIP model is effective and can improve over the CLIP pipeline in zero-shot learning tasks, its performance can be further enhanced through finetuning on a suitably diverse dataset. This enhancement is critical for tasks requiring nuanced understanding of visual and textual data. Future work could explore the impact of finetuning on other datasets or using different finetuning strategies to further understand the adaptability of the ComCLIP model.

\begin{table}[t]
  \caption{Comparison of accuracy (\%) on ComVG, and SVO-Probes using ComCLIP and finetuned ComCLIP. }
  \centering
 \resizebox{\columnwidth}{!}{
  \begin{tabular}{lcccccccc}
    \toprule
    \multirow{2}*{Method} & \multicolumn{4}{c}{ ComVG} & \multicolumn{4}{c}{SVO-Probes}\\
    \cmidrule(lr){2-5} \cmidrule(lr){6-9}
    & {Sub} &{Pred} &{Obj}& Ave&{Sub} &{Pred} &{Obj}&Ave \\
       \midrule 
       CLIP &88.61 & 68.52 & 93.85&83.66 & 85.53 & 80.77 & 90.53 &85.61\\
   ComCLIP  &90.04 & 69.06 & 94.78 &84.63 & 86.70 & 81.87 & 90.67 &86.41\\
   Finetuned ComCLIP  &\textbf{92.14} & \textbf{69.74} & \textbf{97.83} &\textbf{86.98} & \textbf{87.44} & \textbf{81.90} & \textbf{92.48} &\textbf{87.99}\\
  \bottomrule
  \end{tabular}
  }
 \label{tab:finetunecomclip}
\end{table}

\section{Instance-level Image-text Matching Baselines}
We further evaluate ComCLIP's applicability to instance-level image-text matching models by integrating it with SGRAF~\cite{Diao2021SGRAF} on the ComVG dataset. This implementation involves processing the input texts with the same parsing technique used in ComCLIP, coupled with the utilization of grounded image regions for computing the matching score, followed by a reweighting step. The integration of ComCLIP results in a notable performance enhancement: the matching accuracy increases from 76.79\% without ComCLIP to 78.9\% with ComCLIP.

\section{Detailed Algorithm}

The detailed ComCLIP algorithm is summarized in Algorithm~\ref{algo:algo}.



\end{document}